%% file: main.tex
\documentclass[final]{cvpr}

\usepackage{times}
\usepackage{epsfig}
\usepackage{graphicx}
\usepackage{amsmath}
\usepackage{amssymb}

\usepackage{comment}
\usepackage{tablefootnote}
\usepackage[dvipsnames]{xcolor}
\usepackage{bm}
\usepackage{float}
\usepackage[pagebackref=true,breaklinks=true,colorlinks,bookmarks=false]{hyperref}
\usepackage[ruled,vlined,linesnumbered]{algorithm2e}
\usepackage[resetlabels]{multibib}

\SetArgSty{textnormal}
\SetKwInput{KwDef}{Define}

\newcommand{\ci}[1]{\scriptsize $\pm #1$}
\newcommand{\best}[1]{{\boldmath\bfseries#1}} 
\newcommand{\sbest}[1]{\underline{#1}} 
\newcites{New}{Appendix References}

\begin{document}

\title{Using Shape to Categorize: Low-Shot Learning with an Explicit Shape Bias}

\author{Stefan Stojanov, Anh Thai, James M. Rehg\\
Georgia Institute of Technology\\
{\tt\small \{sstojanov, athai6, rehg\}@gatech.edu}}

\maketitle

\input{sections/00_abstract.tex}

\input{sections/0_introduction.tex}
\input{sections/1_related_work}

\input{sections/2_approach}

\input{sections/3_experiments}
\input{sections/4_conclusion}
\section{Acknowledgement}
We would like to thank Rohit Gajawada and Jiayuan Chen for their help with initial data collection. We also thank Zixuan Huang and Miao Liu for their helpful discussion of the paper draft. This work was supported by NSF award 1936970 and NIH award R01-MH114999.


{\small
\bibliographystyle{ieee_fullname}
\bibliography{egbib}
}

\clearpage
\appendix
\input{sections/appendix.tex}

\clearpage
{\small
\bibliographystyleNew{ieee_fullname}
\bibliographyNew{egbib_supp}
}

\end{document}

%% file: sections/00_abstract.tex
\begin{abstract}
It is widely accepted that reasoning about object shape is important for object recognition. However, the most powerful object recognition methods today do not explicitly make use of object shape during learning. In this work, motivated by recent developments in low-shot learning, findings in developmental psychology, and the increased use of synthetic data in computer vision research, we investigate how reasoning about 3D shape can be used to improve low-shot learning methods' generalization performance. We propose a new way to improve existing low-shot learning approaches by learning a discriminative embedding space using 3D object shape, and using this embedding by learning how to map images into it. Our new approach improves the performance of image-only low-shot learning approaches on multiple datasets. We also introduce Toys4K, a 3D object dataset with the largest number of object categories currently available, which supports low-shot learning. \footnote{The code and data for this paper are available at our project page \href{https://rehg-lab.github.io/publication-pages/lowshot-shapebias/}{https://rehg-lab.github.io/publication-pages/lowshot-shapebias/}}
\end{abstract}
\vspace{-15pt}

%% file: sections/0_introduction.tex
\section{Introduction}

Understanding the role of 3D object shape in categorizing objects from images is a classical topic in computer vision~\cite{Mundy2006, edelman1999representation, Ullman1996}, and the early history of object recognition was dominated by considerations of object shape. For example, David Marr's influential theory~\cite{marr2010vision} posits that image-based recognition should be formulated as a sequence of information extraction steps culminating in a 3D representation to be used for recognition. The difficulty of reliably extracting 3D shape from images, combined with the availability of large-scale image datasets~\cite{deng2009imagenet, krizhevsky2009learning}, motivated the modern development of purely appearance-based approaches to recognition and categorization. This has culminated in current approaches such as CNNs that learn feature representations directly from images. Moreover, a study by Geirhos et al.~\cite{geirhos2018imagenet} of the inductive biases of CNNs trained on ImageNet suggests that categorization performance is driven primarily by a bias towards image texture rather than object shape.\footnote{This study does not speak to the possibility of whether shape could be used more effectively, and it is unclear how much of the bias stems from the composition of the ImageNet dataset itself.}

However, studies of infant learning~\cite{landau1988importance, diesendruck2003specific, landau1998object, gershkoff2004shape} suggest that shape does play a significant role in the ability to rapidly learn object categories from a small number of examples, a task which is analogous to few-shot learning. Both young children and adults who are forced to categorize novel objects based on a few examples display a \emph{shape bias}, meaning that shape cues seem to play a dominant role in comparison to color and texture when inferring category membership. These studies beg the question of whether information about 3D object shape could be useful in learning to perform few-shot categorization from images. While prior work has demonstrated effective approaches to object categorization using 3D shapes as input~\cite{qi2017pointnet, qi2017pointnet++, wang2019dynamic, wu20153d, chenpointmixup}, and there is a large literature on few-shot learning from images alone~\cite{snell2017prototypical, vinyals2016matching, hu2018relation, rusu2018meta, Ye_2020_CVPR, tian2020rethink, finn2017model}, the question of how shape cues could be used to learn effective representations for image-based low-shot categorization has not been investigated previously.

\input{figure_tex/approach_fig}

The goal of this paper is to explore the incorporation of a shape bias in SOTA approaches to few-shot object categorization and thereby investigate the utility of shape information in category learning. We leverage the recent availability of datasets of 3D object models with category labels, such as ModelNet40 \cite{wu20153d} and ShapeNet \cite{chang2015shapenet}. By sampling surface point clouds and rendering images of these models, we can construct datasets that combine 3D shape and image cues. Unfortunately, however, ShapeNet and ModelNet contain a relatively small number of object categories (55 and 40 respectively), making it difficult to test categorization at a sufficient scale. To resolve this limitation, we introduce a new 3D object dataset, \emph{Toys4K} consisting of 4,179 3D objects from 105 object categories, designed to contain categories of objects that are commonly encountered by infants and children during their development.

We report on two sets of investigations. First, we examine the relative effectiveness of purely image-based and purely shape-based approaches to few-shot categorization. We demonstrate that purely shape-based few-shot learning outperforms image-based approaches, and establish an empirical upper bound on the effectiveness of a shape bias. Second, we develop a novel approach for training an image embedding representation for low-shot categorization which incorporates an explicit shape bias, which we outline in Figure~\ref{fig:approach_overview}. We benchmark this approach on a representative set of SOTA few-shot learning architectures and demonstrate that the incorporation of shape bias results in increased generalization accuracy over image-based training alone. In summary, this paper makes the following contributions:
\vspace{-0.3em}
\begin{itemize}
    \vspace{-2pt}
    \setlength\itemsep{-0.3em}
    \item A new approach to add explicit shape-bias to existing low-shot image classification methods, utilizing 3D object shape to learn similarity relationships between objects, which leads to improved low-shot performance.
    \item The first evidence that shape information can enable image-based low-shot classifiers to generalize with higher accuracy to novel object categories.
    \item Toys4K - new 3D object dataset containing approximately twice the number of object categories as previous datasets which can be used for low-shot learning.
\end{itemize}
\vspace{-7pt}

%% file: figure_tex/approach_fig.tex
\begin{figure*}[h]
\begin{center}
\includegraphics[width=\linewidth]{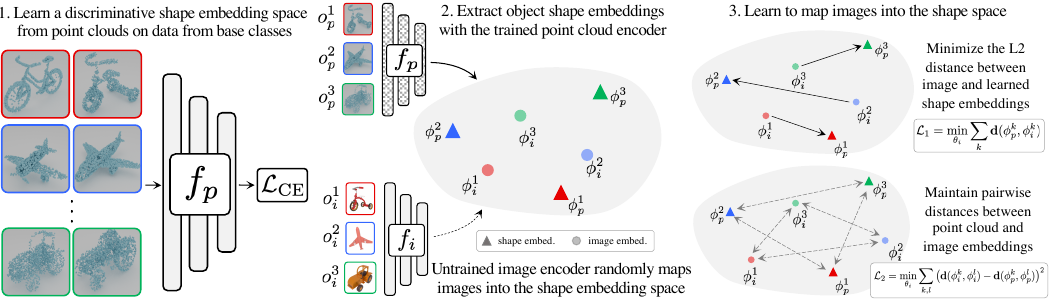}
\end{center}
\vspace{-10pt}
\caption{To perform low-shot learning with shape-bias, we train a embedding space defined by a point cloud encoder $f_p$ trained with cross-entropy. Shape based embedding spaces are more discriminative than image-based ones (see Tbl.~\ref{tbl:appearance_vs_shape}). We extract object shape embeddings, and train an image encoder $f_i$ to map images into the shape space. If trained successfully, $f_i$ will have the discriminative properties of $f_p$.}
\label{fig:approach_overview}
\vspace{-15pt}
\end{figure*}

%% file: sections/1_related_work.tex
\section{Related Work}
\label{sec:related-work}
\textbf{Object Recognition from Synthetic Data}

A large body of work focuses on appearance \cite{su2015multi, monti2017geometric, qi2016volumetric, feng2018gvcnn, ho2020exploit}, point cloud \cite{qi2017pointnet, wang2019dynamic, qi2017pointnet++, chenpointmixup} and voxel \cite{wu20153d, qi2016volumetric} based recognition of synthetic object data with category taxonomies based on object shape such as ModelNet40~\cite{wu20153d}. The trade-offs between learning using point clouds, depth maps, voxels, or images have been studied by \cite{su2015multi, qi2016volumetric} but their study focuses on standard supervised classification and does not extend to low-shot classification of novel object categories or on combining shape and appearance information during learning.

\textbf{Low-Shot Learning}\\
Low-shot learning algorithms can be categorized into two broad sets. Optimization-based algorithms such as MAML \cite{finn2017model, finn2018probabilistic}, LEO \cite{rusu2018meta}, and Reptile \cite{nichol2018reptile}, which during the base-classes training stage, attempt to learn a representation that can quickly be adapted using small amounts of information with gradient-based learning in the low-shot learning stage. Metric learning-based methods such as Prototypical \cite{snell2017prototypical}, Matching \cite{vinyals2016matching}, and Relation \cite{sung2018learning} networks, as well as the more recent SimpleShot~\cite{wang2019simpleshot}, FEAT~\cite{Ye_2020_CVPR}, and RFS~\cite{tian2020rethink}  aim to use the base class data to learn a similarity metric that will also be discriminative for novel classes during the low-shot phase. Despite their simplicity, metric-based approaches have superior performance on low-shot learning benchmarks~\cite{wang2019simpleshot, Ye_2020_CVPR}. Our approach of adding shape bias belongs to the latter category, and compared to both is the first approach to combine both appearance and shape information for low-shot learning.

\textbf{3D Object Shape Datasets} \\
Other related works focus on building datasets of 3D object models for recognition, single image object shape reconstruction and shape segmentation \cite{shilane2004princeton, tatsuma2012large, zhou2016thingi10k, koch2019abc, wu20153d, chang2015shapenet, stojanov2019incremental}. The most widely used 3D shape datasets with category labels are ModelNet40 \cite{wu20153d} with 12K object instances of 40 categories with no object surface material properties, ShapeNetCore.v2 \cite{chang2015shapenet} with 52K objects of 55 object categories with basic surface texture properties (basic shading and UV mapping, but no physically based materials). The ShapeNetSem split of ShapeNet consists of over 100 categories but is unsuitable for recognition since individual object instances are assigned to multiple categories. Datasets such as ABC \cite{koch2019abc} and Thingi10k \cite{zhou2016thingi10k} claim higher mesh quality than previous datasets but lack object category annotation, making them more suitable for low-level tasks like surface normal estimation and category agnostic shape reconstruction. The ModelNet40 and ShapeNet datasets were scraped from online repositories and have categories largely based on the data that was available in these repositories. In contrast, our new Toys4K is curated specifically for testing the generalization ability of learned representations to new classes. Compared to the aforementioned datasets, Toys4K consists of highly diverse object instances within a category (evident in Figure~\ref{fig:toys45K_viz}, detailed composition is included in the supplement) and has the highest number of individual object categories despite its smaller total size. 

\textbf{Multi-modal Learning}\\
Aligning representations from different data modalities has been extensively studied in vision and language works on zero-shot learning \cite{xian2018feature, hubert2017learning, schonfeld2019generalized, frome2013devise}. More recently, Schwartz et al.~\cite{schwartz2019baby} and Xing et al.~\cite{xing2019adaptive} improve low shot image classification performance on standard low-shot datasets by combining the representation learned through the appearance modality (images) with language model word vector embeddings. In comparison, we combine appearance (images) and shape (point clouds) to learn a representation for low shot object recognition that is biased to object shape and leads to better low-shot generalization. It is important to note that these works use multi-modal information for the low-shot queries at test time, whereas our approach only uses multi-modal information for the low-shot support set.

Another category of multi-modal learning works focuses on learning joint embedding spaces of 3D meshes and images for image-based 3D shape retrieval \cite{lee2018cross, li2015joint}. While these works focus on retrieval for the same object categories at training and testing time, our work focuses on combining appearance and shape information for low-shot generalization to \emph{novel} object categories.

%% file: sections/2_approach.tex
\section{Using Shape for Low-Shot Classification}
\label{sec:approach}

In principle, 3D shape is an attractive representation for object recognition~\cite{marr2010vision, osada2002shape, li2015joint, lee2018cross} due to its invariance to the effects of viewpoint, illumination, and background, which can be challenging for appearance-based approaches. While appearance-based methods may be able to model these sources of variation given sufficient training images, there is always a question of how well such models can generalize to novel categories and objects~\cite{geirhos2018imagenet}.

Despite its potential advantages, no previous work on low-shot learning has utilized 3D shape, for at least two reasons: 1) It is unclear how to leverage 3D shape in improving \emph{image-based} low shot learning;\footnote{Our focus is on few-shot methods in which the queries are \emph{images}, with no 3D shape information available, as this is the most general and useful paradigm.} 2) There is a lack of 3D shape datasets that contain a sufficient number of object categories to support effective experimentation. This is due to the additional data requirements of few-shot learning: The training/validation/testing split is over different classes and not data points of the same class~\cite{ren2018meta, vinyals2016matching} in order to effectively test generalization to unseen classes. 

To explain this issue more formally, let $\mathcal{D}^{\text{train}}$ denote the base classes, and $\mathcal{D}^{\text{val}}$ and $\mathcal{D}^{\text{test}}$ denote the validation and testing sets, respectively, where these sets comprise a disjoint partition of the total available classes. The base classes must be sufficiently large and diverse to learn an effective feature representation in the training phase, and the $\mathcal{D}^{\text{val}}$ set must similarly support the accurate assessment of low-shot generalization ability during hyperparameter tuning (i.e. model selection while training on the base classes). The $\mathcal{D}^{\text{test}}$ set is used to generate labeled low-shot training examples (supports), and unlabelled low-shot testing examples (queries), which are used to evaluate the generalization performance of the model at testing time, which we refer to as the \emph{low-shot phase}. As a result of these constraints, the standard 3D shape datasets ModelNet40~\cite{wu20153d} and ShapeNet55~\cite{chang2015shapenet} can only support 10-way and 20-way testing, respectively. If the number of testing classes is insufficient, the estimation of the generalization performance of the method may be inaccurate.

In this section, we describe our two primary contributions which address the limitations described above. In \S~\ref{subsec:algorithm} we present our novel method for introducing \emph{shape bias} in learning a low-shot image representation. In \S~\ref{subsec:dataset}, we introduce a novel 3D object category dataset, \emph{Toys4K}, consisting of 4,179 object instances organized into 105 categories, with an average of 35 objects (3D meshes) per category. Toys4K supports up to 50-way classification, expanding well beyond ModelNet40 and ShapeNet55 (see Fig.~\ref{fig:dataset-accuracy}). 

\input{figure_tex/combined_fig_2}

\subsection{Low-Shot Learning with Shape Bias}
\label{subsec:algorithm}

We begin by describing the \emph{problem formulation:} We assume that shape data in the form of 3D point clouds is available for each RGB image in a dataset. We achieve this by rendering RGB images from the 3D models. 3D shape information is used directly during training and validation, in order to construct a representation with an explicit shape bias. In addition, during the low-shot phase, episodes are generated so that point clouds are available for the support objects, but \emph{not} for the query objects. This assumption allows for both appearance and shape information to be used in building class prototypes, but \emph{inference is done using images only}. The distinction between image only low-shot learning and our new setting is illustrated in Figure~\ref{fig:setting}. 

In this work, we adopt a low-shot learning approach based on a \emph{metric embedding space}. In this approach, $\mathcal{D}^{\text{train}}$ is used to learn a function $f_i$ that maps the input data into an embedding space where object instances of the same category are close and instances of different categories are far apart,  according to some distance metric. This mapping can be fixed after being learned from $\mathcal{D}^{\text{train}}$ or fine-tuned further, depending upon the algorithm design. During the low-shot phase, the supports and queries are mapped into the embedding space (see Figure~\ref{fig:setting}), and the queries are classified according to a nearest neighbor or nearest class prototype (e.g. support centroid) rule. Metric-based low-shot learning has high accuracy \cite{wang2019simpleshot} and is significantly more computationally efficient than approaches that fine-tune on the low-shot supports. We first demonstrate that shape-based low-shot learning allows for better generalization than image-based low-shot learning, and then show how a shape-based embedding with high generalization ability can be used to improve image-based low-shot classification.

\noindent\textbf{Shape-based low-shot learning outperforms image-based low-shot learning}

We perform a simple empirical study to determine whether shape has an advantage for low-shot generalization. We train two embedding spaces, one using image data and one using point cloud data. For each type of data, we follow the SimpleShot \cite{wang2019simpleshot} approach, meaning that we train a classifier using cross-entropy on $\mathcal{D}^{\text{train}}$ and use the learned feature space (output of the last pooling layer) to perform nearest centroid-based low-shot classification in normalized Euclidean space. We use a ResNet18 \cite{he2016deep} for image learning  and a DGCNN \cite{wang2019dynamic} for point cloud learning on the ModelNet40-LS dataset (see \S~\ref{sec:experiments}). 

We present the results in Tbl.~\ref{tbl:appearance_vs_shape}, and as might be expected, see significantly higher low-shot performance for the point cloud model relative to the image model. This quantifies the improvement in generalization to novel categories as as result of using a 3D shape-based representation and suggests that 3D shape can yield a more discriminative embedding space. The question then is \emph{how can this benefit be retained when testing the model on image data alone?}

\noindent\textbf{Combining Appearance and Shape}

Figure~\ref{fig:approach_overview} illustrates our approach to using the 3D shape information available at training time in order to learn how to embed the image-only queries. First, we train a low-shot point-cloud based classifier on the set of base-classes $\mathcal{D}^{\text{train}}$, resulting in an a highly discriminative embedding space for both seen and novel categories. We then extract point cloud embeddings for each object in the training set and train a CNN to map images into the shape embedding space.

Let $\mathcal{D}$ be a dataset of paired object point clouds $o_p$ and images $o_i$, partitioned into 
$\mathcal{D}^{\text{train}}$, $\mathcal{D}^{\text{val}}$, and $\mathcal{D}^{\text{test}}$. Let $f_{p}(x)\colon N\times \mathbb{R}^3 \to \mathbb{R}^d$ denote the trained function for mapping point clouds of size $N$ into an embedding space of dimension $d$.
This embedding space is optimized to yield favorable metric properties for low shot classification,
using the labelled point cloud data in $\mathcal{D}^{\text{train}}$. Our goal is then to learn a second mapping, $f_i(x)\colon \mathbb{R}^{H\times W\times 3}\to \mathbb{R}^d$, where $H,W$ are the image height and width, from images into the shape embedding space defined by $f_{p}(x)$. We denote point cloud embeddings as $f_{p}(o_p) = \phi_p$ and image embeddings as $f_i(o_i) = \phi_i$.

\input{figure_tex/combined_fig_1}

We train a model that learns the mapping from images to shape embeddings by minimizing two loss functions (see part 3 of Figure~\ref{fig:approach_overview}). For a mini-batch $\mathcal{B} \subset \mathcal{D}^{\text{train}}$ the first loss minimizes the squared Euclidean distance (which we denote as $\mathbf{d}(x,y)$) between the learned point cloud embeddings, and the image based embeddings
\[\mathcal{L}_1 = \sum_{(o_i, o_p) \in \mathcal{B}} \mathbf{d}\big(\phi_i, \phi_p\big). \]
The second loss constrains the pairwise distances between the image embeddings of different object instances to be the same as the pairwise distances of the learned shape embeddings. Let $\mathcal{I}$ denote the set of all $(k,l) = \big((o_i^k, o_p^k), (o_i^l, o_p^l)\big)$ object instance data pairs in a mini-batch. We define the second loss as
\[\mathcal{L}_2 = \sum_{(k,l) \in \mathcal{I}} \big(\mathbf{d}(\phi_p^k, \phi_p^l) - \mathbf{d}(\phi_i^k, \phi_i^l)\big)^2. \]
During training, both losses are minimized with equal weight. Validation for choosing $f_i$ is done by nearest centroid classification on $\mathcal{D}^{\text{val}}$. In Section~\ref{sec:experiments} we show that minimizing only $\mathcal{L}_1$ results in convergence without learning to match the distribution of the shape embedding well on the training set, resulting in poor performance.

\noindent\textbf{Inference:} During the low-shot phase, as shown in Figure~\ref{fig:setting}, class prototypes are built by averaging the shape $\phi_{p}$ and image $\phi_i$ embeddings for each support object, whereas only image information is used to map the query objects via $f_i$. The queries are classified based on the nearest centroid to the query embedding. This inference procedure is used for all algorithms in this paper that combine both image and shape information, with the exception of FEAT~\cite{Ye_2020_CVPR}, which uses an additional set-to-set mapping. 

It is important to understand how the shape-biased encoder performs when there is no explicit shape information available in the low-shot phase, and what is the gain in accuracy by making shape available for building class prototypes. To this end, in \S~\ref{sec:shape_bias_improvements} we also evaluate the setting where there are no point clouds available in the low-shot phase.

\noindent\textbf{Why is mapping images to shape embeddings difficult?}
If the mapping $f_i(x)$ is learned exactly, it would map images to their corresponding point cloud embeddings so that
\[\forall (o_i, o_p) \in \{\mathcal{D}^{\text{train}} \cup \mathcal{D}^{\text{val}} \cup \mathcal{D}^{\text{test}}\}, ||\phi_i - \phi_p||_2 =0. \]
This is challenging, however, since $f_i$ can only be trained on the base classes in $\mathcal{D}^{\text{train}}$, requiring it to correctly extrapolate to the metric properties of objects from novel classes.

We perform a simple test to validate the feasibility of mapping images to shape embeddings in general and establish an empirical upper bound. We perform this by simply minimizing the $L_2$ distance between the images and their corresponding shape embeddings on combined data from base classes, validation and test classes ($ \mathcal{D}^{\text{train}} = \{\mathcal{D}^{\text{train}} \cup \mathcal{D}^{\text{val}} \cup \mathcal{D}^{\text{test}}\}$). This model is referred to as Image + Point Cloud Oracle in Table~\ref{tbl:appearance_vs_shape} and provides empirical evidence that it is possible to learn how to map images into a shape embedding space with high accuracy when all of the data is available. This model's performance closely matches that of the shape-only model, and significantly outperforms the image-based approach, providing further evidence that extrapolating the metric properties of the shape-embedding space to novel categories is the key challenge in learning to map images to shape embeddings.

\subsection{Toys4K Dataset}
\label{subsec:dataset}
An object dataset with a high number of diverse categories and high-quality 3D meshes is essential to study whether leveraging 3D object shape can enable improved low-shot generalization. We satisfy this requirement with our new \emph{Toys4K dataset}. While it is possible to use existing datasets such as ModelNet40 and ShapeNet (which we include in our experiments), the limited number of categories is an obstacle to few-shot learning. For example, applying standard training/validation/test ratios (e.g. from mini-ImageNet~\cite{vinyals2016matching}) to the 40 categories in ModelNet40 results in a 20-10-10 split, which limits the possibilities for many-way testing. A comparison of Toys4K to prior datasets is available in Table~\ref{tbl:dataset-comparison}. In Figure~\ref{fig:dataset-accuracy} we demonstrate that many-way low-shot classification on Toys4K is a challenging task in comparison to ModelNet40 and ShapeNet.

Toys4K consists of 4,179 object instances in 105 categories, with an average of 35 object instances per category with no less than 15 instances per category, allowing for 5 support 10 query low-shot episodes to be formed. Fig.~\ref{fig:toys45K_viz} provides an example of the quality and variety of the models. Further details on the dataset composition are available in the supplement. Toys4K was collected by selecting freely-available objects from Blendswap~\cite{blendswap}, Sketchfab~\cite{sketchfab}, Poly~\cite{poly} and Turbosquid~\cite{turbosquid} under Creative Commons and royalty-free licenses. Our list of object categories was developed in collaboration with experts in developmental psychology to include categories of objects available and relevant to children in their infancy. We manually selected each object and manually aligned the objects within each category to a canonical coordinate system that is consistent across all instances in that category.

%% file: figure_tex/combined_fig_2.tex
\begin{figure*}
\centering
\begin{minipage}{.61\linewidth}
  \centering
  \includegraphics[width=\linewidth]{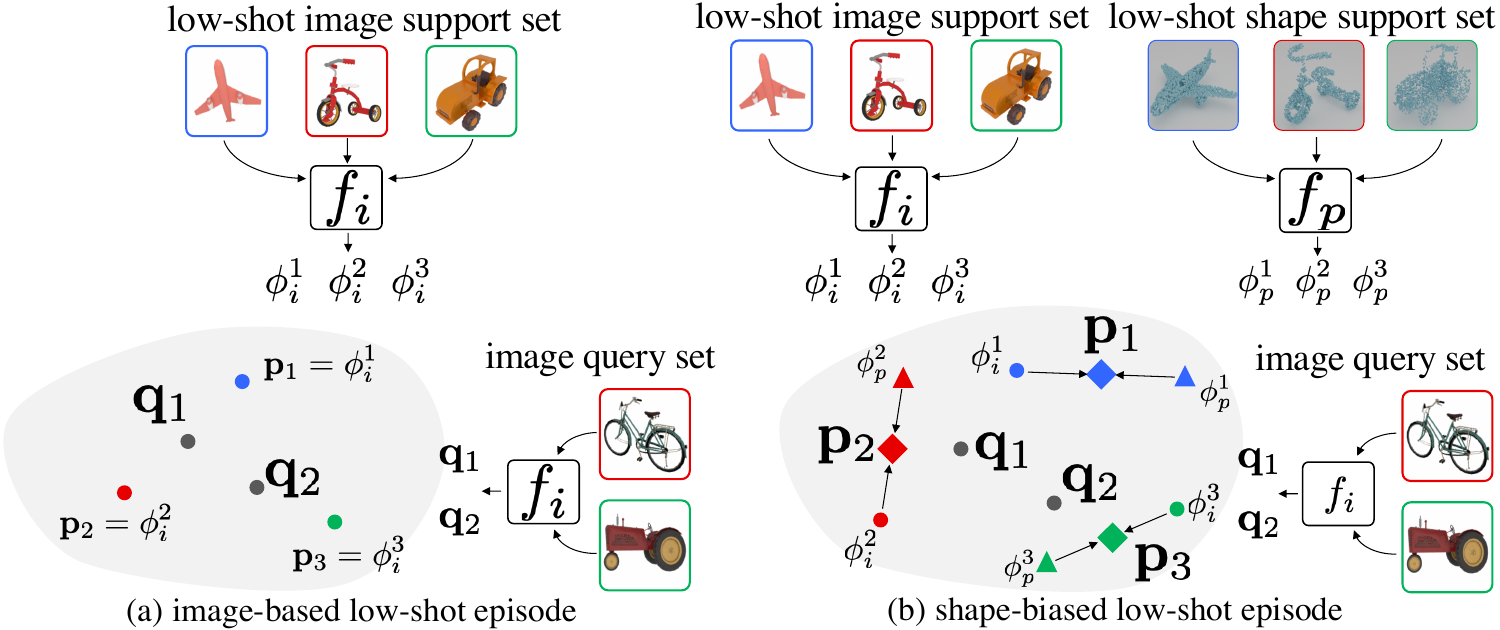}
  \caption{(a) The standard setting: Prototypes are formed from images. (b) Our novel shape-biased setting: Image and shape embeddings are averaged. In both cases, the image-only queries $q_i$ can be classified 
  by identifying the closest prototype $p_j$. The training process for the mapping functions $f_i$ and $f_p$ is illustrated in Figure~\ref{fig:approach_overview}.}
  \label{fig:setting}
\end{minipage}
\hspace{5pt}
\begin{minipage}[t!]{.33\linewidth}
\begin{table}[H]
\scalebox{0.8}{
\renewcommand{\arraystretch}{1.15}
\centering
\begin{tabular}{c|cc|cc|}
 & \multicolumn{2}{c|}{5-way} & \multicolumn{2}{c|}{10-way}  \\
\cline{2-5}
& 1-shot & 5-shot &  1-shot & 5-shot \\
\hline
Image       & 58.99 & 74.29 & 45.82 & 62.73 \\    
Point Cloud & 66.02 & 83.61 & 54.44 & 75.26 \\
\hline
\begin{tabular}{@{}c@{}}Img + Ptcld \\ Oracle \end{tabular}   &  68.04 & 82.07 & 57.03 & 73.11
\end{tabular}
}
\vspace{3pt}
\caption{On ModelNet40-LS, low-shot generalization is higher for point-cloud based learning than image based learning, justifying our approach in combining the modalities. Oracle model has access to both image and point cloud information. See text for details.}
\label{tbl:appearance_vs_shape}
\end{table}
\end{minipage}
\vspace{-15pt}
\end{figure*}

%% file: figure_tex/combined_fig_1.tex
\begin{figure*}
\centering
\begin{minipage}{.65\linewidth}
\begin{center}
\includegraphics[width=\linewidth]{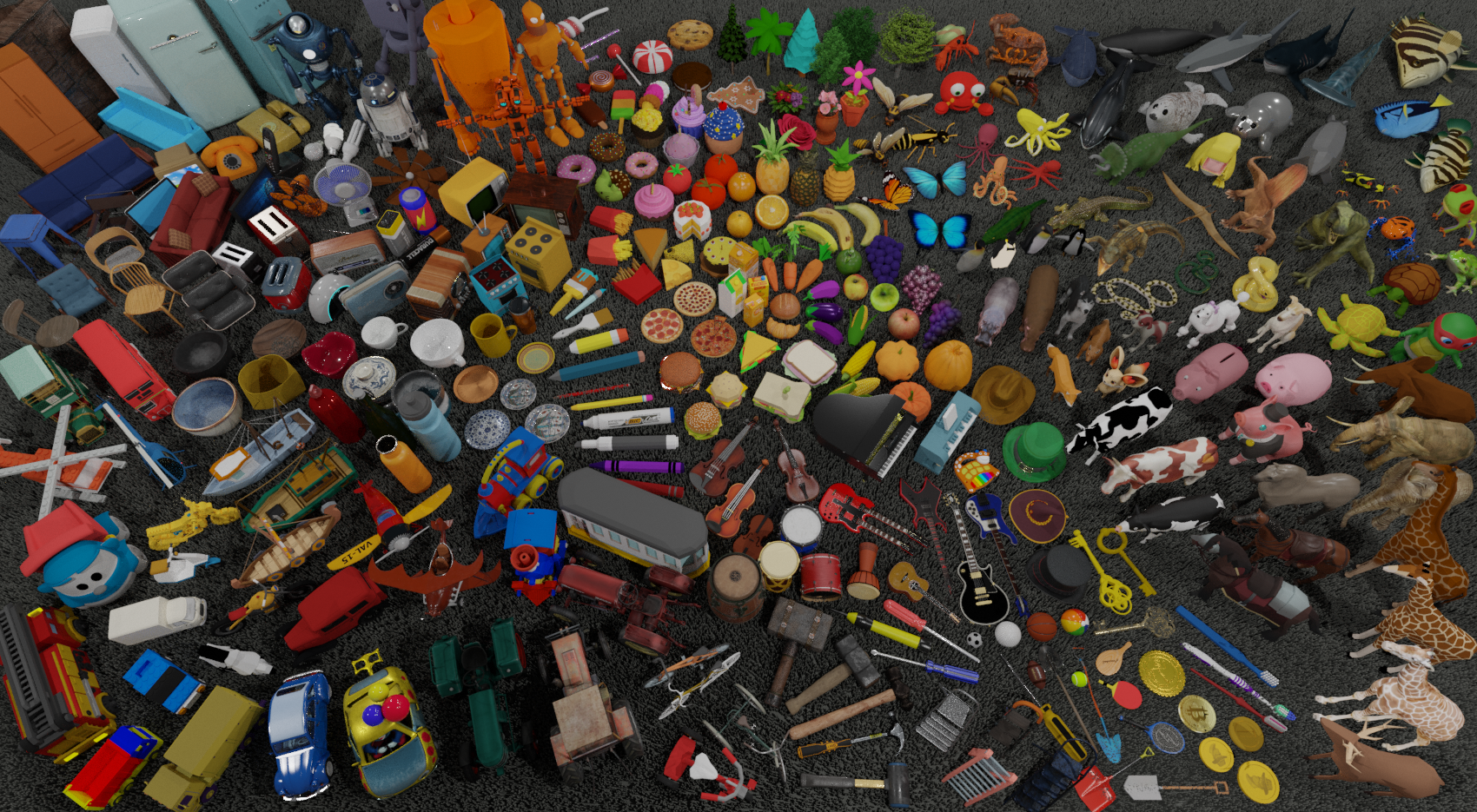}
\end{center}
\vspace{-10pt}
\caption{Approximately one third of the objects in Toys4K, a new dataset of 3D assets for low-shot object learning using object appearance and shape information.}
\label{fig:toys45K_viz}
\end{minipage}
\hfill
\begin{minipage}[t!]{.33\linewidth}
\begin{center}
\includegraphics[width=\linewidth]{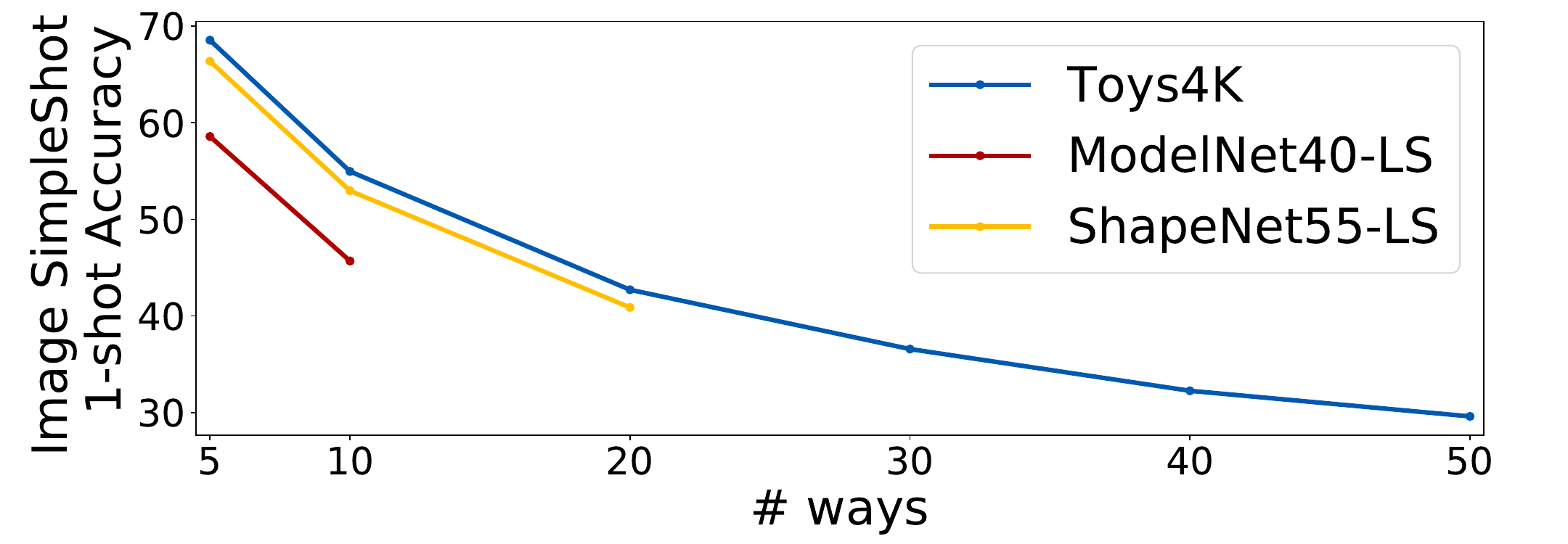}
\end{center}
\vspace{-10pt}
\caption{The high number of categories in Toys makes low-shot learning on Toys4K a challenging task.}
\label{fig:dataset-accuracy}

\begin{table}[H]
\scalebox{0.8}{
\begin{tabular}{l|c|c}
\textbf{Dataset} & \textbf{Instances} & \textbf{Categories} \\ 
\hline
Toys4K                              & 4,179     & 105 \\    
ModelNet40~\cite{wu20153d}          & 12,311    & 40  \\
ShapeNet~\cite{chang2015shapenet}   & ~52,000   & 55  \\
Thingi10K~\cite{zhou2016thingi10k}  & ~10,000   & N/A \\
ABC~\cite{koch2019abc}              & 750K      & N/A \\

\end{tabular}
}
\vspace{5pt}
\caption{Toys4K has the most categories of any available dataset of 3D objects.}
\label{tbl:dataset-comparison}
\end{table}

\end{minipage}
\vspace{-5pt}
\end{figure*}

%% file: sections/3_experiments.tex
\section{Experiments}
\label{sec:experiments}
In this section, we perform an empirical evaluation of the benefit of explicit shape bias on multiple datasets and image-only low-shot learning algorithms.
\subsection{Datasets}
In addition to our new dataset, Toys4K, we use the 3D object category datasets ModelNet40~\cite{wu20153d}, ShapeNet~\cite{chang2015shapenet}. For descriptions of the datasets please refer to \S~\ref{sec:related-work}. We render images using the Cycles ray tracing renderer in Blender~\cite{blender} using uniform lighting on white backgrounds. For all datasets, camera pose is randomly sampled for 25 views of each object with azimuth $\psi \in [0, 360]$ and elevation $\theta \in [-50,50]$ degrees. Object surface point clouds are sampled from the 3D object meshes.

\textbf{Toys4K} is our new low-shot learning dataset is described in detail in \S~\ref{subsec:dataset}. We use a split of 40, 10, 55 for base, low-shot validation, and testing classes, respectively. For Toys and all other datasets, the split is designed such that the categories with most classes are in the training set, and the validation and testing classes are randomly chosen from the remainder of the data. 

\textbf{ModelNet40-LS} is the existing ModelNet40~\cite{wu20153d} dataset, with a 20, 10, 10 split for base, low-shot validation and testing classes respectively.

\textbf{ShapeNet-LS} is the existing ShapeNetCore.v2~\cite{wu20153d} dataset, with a 25, 10,  20 split for base, low-shot validation and testing classes respectively, using a reduced subset of object samples per category to reduce training time due to the high data imbalance.

\subsection{Baselines}
Regarding low-shot learning, we compare with the classical low-shot learning method Prototypical Networks~\cite{snell2017prototypical}, and the state-of-the-art algorithms FEAT~\cite{Ye_2020_CVPR}, RFS~\cite{tian2020rethink}, and SimpleShot~\cite{wang2019simpleshot}. With respect to learning joint embeddings, we compare with a simple triplet loss-based approach that learns joint embeddings of images and point clouds. All baselines use a standard ResNet18~\cite{he2016deep} as a backbone for image encoding and a DGCNN \cite{wang2019dynamic} to encode point clouds. In the supplement we perform an ablation study over different point cloud architectures including PointNet~\cite{qi2017pointnet} and PointNet++~\cite{qi2017pointnet++}. Our low-shot learning baseline implementations were all validated by re-creating the results from the original papers.\footnote{Experiment implementation details included in the supplement}

\textbf{SimpleShot}~\cite{wang2019simpleshot} is a simple low-shot learning baseline algorithm that outperforms many recent methods. It makes use of an embedding space learned by a CNN by training on the base training classes for a standard classification task using cross-entropy loss. Validation and testing are done using a nearest neighbor classifier in the learned embedding space, with feature normalization and training set mean subtraction resulting in improved performance.

\textbf{RFS}~\cite{tian2020rethink} is another simple low-shot learning algorithm that is competitive with many recent approaches. Training the embedding space is done using cross-entropy on the training set, but at testing time, a simple logistic regression classifier is learned for each low-shot episode. In the original work, the authors show that training a set of embedding models with distillation slightly improves performance. We omit this for a fair comparison with all metric-based works since this addition would likely lead to performance improvements across the board.

\input{table_tex/modelnet_res_table}
\input{table_tex/toys_res_table}
\input{table_tex/shapenet_res_table}

\textbf{Prototypical Networks}~\cite{snell2017prototypical} is a standard metric-based low shot learning approach, which uses the base class set to create low-shot episodes and learn a feature space that embeds object instances close or far based on visual similarity.

\textbf{FEAT}~\cite{Ye_2020_CVPR} builds on Prototypical Networks by learning an additional set-to-set function implemented as a Transformer~\cite{vaswani2017attention} on top of a cross-entropy pre-trained embedding space to refine the class prototypes used for low-shot classification. FEAT achieves state of the art performance for inductive low-shot learning. Note that FEAT requires separate retraining for each $n$-way $m$-shot configuration \footnote{Since none of the datasets have more than 10 classes for validation, the 20 and 30-way evaluations are done using a model trained for 10-way classification.}.

\textbf{Triplet} We use a simple triplet loss-based approach as a baseline algorithm with access to both image and shape information during training, similar to prior approaches in shape retrieval \cite{lee2018cross}. A joint embedding is learned by using triplet loss \cite{chechik2009large, schroff2015facenet}, creating positive pairs between image and point cloud features of same objects, and negative pairs between image and shape features from different object instances. Empirically we found that this performs better than using category labels. Inference is done by nearest centroid classification, building class prototypes that contain both appearance and shape information by averaging the individual support features.

\subsection{Explicit Shape-Bias Improves Image-Based Generalization}
\label{sec:shape_bias_improvements}
We evaluate our method of adding shape bias to low-shot learning algorithms with state of the art low-shot image-only classification algorithms and show that shape bias improves performance in a low data regime. We present results on multiple datasets in Tables~\ref{tbl:modelnet-res-table},~\ref{tbl:toys-res-table}, and~\ref{tbl:shapenet-res-table} where we refer to models as Shape Bias (w/pc) if the shape-biased image encoder uses point cloud information to build prototypes (see Fig~\ref{fig:setting}(b)) and (wo/pc) if there are no point clouds used to build prototypes for both validation and testing (see Fig~\ref{fig:setting}(a)). Our approach of introducing shape bias, when trained with $\mathcal{L}_1$ and $\mathcal{L}_2$ losses improves the performance of image-only low-shot recognition algorithms in the low-data, one-shot learning regime for the SimpleShot and FEAT algorithms by up to 6\%-points. For the (w/o pc) models that do not have any explicit shape information in the low-shot phase, we see a smaller one-shot improvement, but good five-shot performance. This indicates that shape bias is useful \emph{without} any explicit shape information in the low-shot phase, and suggests possible future improvements by using strategies other than averaging to combine image and shape information in the low-shot phase.
\input{figure_tex/pairwise_loss_analysis}

We add shape-bias to SimpleShot by directly using the learned image to shape-mapping $f_i$ for nearest class mean classification, whereas for FEAT we train the set-to-set Transformer module on top of $f_i$, fine-tuning the model end-to-end as in the original FEAT design. The object shape embeddings for the low-shot supports are fixed and not trained further. Notice that as the total number of categories (the number of low-shot ways) increases, \emph{the improvement in one-shot performance increases.} Further, our approach of learning shape bias significantly outperforms the triplet-loss based approach, indicating that first learning an embedding space with point clouds only is a better strategy than joint training with images and point clouds. All experiments for SimpleShot are averaged over 5 runs and for FEAT are averaged over 3 runs, indicating consistent performance improvements. To ensure statistical significance, for all experiments we perform 5K low-shot episodes and report results with 95\% confidence intervals.\footnote{For further qualitative and quantitative analysis please refer to the supplement.}

\subsection{Analysis of Pairwise Loss}
We perform an analysis to determine the benefit of including the pairwise distance loss $\mathcal{L}_2$. In Figure~\ref{fig:pairwise-analysis}, we plot the pairwise interclass distances of object instances from categories in the validation set for the learned mapping $f_i$ trained either with one loss or both losses (blue and orange respectively), along with the interclass distances in the point cloud embedding that $f_i$ is trained to learn. The greater the overall interclass distance, the better, and ideally the pairwise distance distributions are the same between the learned mapping and the point cloud mapping. Just optimizing $\mathcal{L}_1$ results in learning a poor mapping on both the training set and the novel classes in the validation set, whereas adding the pairwise term $\mathcal{L}_2$ leads to a better approximation of the point cloud embedding. The utility of $\mathcal{L}_2$ is also shown in Table~\ref{tbl:modelnet-res-table}, with the significant improvement over just $\mathcal{L}_1$ on SimpleShot with shape bias.

\subsection{Shape Bias and Failure Analysis}
To better understand the distinctions between the purely image-based low-shot classifier and the shape-biased low shot classifier, we compute the Pearson correlation ($p<0.05$) between the accuracy achieved on the same 5K low-shot episodes for the point cloud model and the shape-biased and image-only classifiers (Figure~\ref{fig:correlation-analysis}). The shape-biased low-shot classifier correlates more strongly with the point cloud model across multiple datasets. This is evidence for a qualitative difference beyond classification accuracy between the shape biased and purely image low-shot classifiers. This would not be possible if the image data was such that it could not be classified differently as a result of introducing shape bias. Furthermore, in Table~\ref{tbl:misclassification-analysis} we see that shape-biased SimpleShot misclassifies similarly to the point cloud SimpleShot, and that there is significant room for improvement by learning to map images into shape embeddings more accurately.
\input{figure_tex/accuracy_correlation_fig}
\input{table_tex/percent_misclassified}

%% file: table_tex/modelnet_res_table.tex
\begin{table*}[h]
\renewcommand{\arraystretch}{1.0}
\begin{center}
\scalebox{0.85}{
\begin{tabular}{l|cc|cc}

Episode Setup $\to$ & 1-shot 5-way  & 5-shot 5-way&  1-shot 10-way& 5-shot 10-way\\

\hline

RFS~\cite{tian2020rethink}           & 56.67 \ci{0.30}  & 72.64 \ci{0.26} & 43.79 \ci{0.16} & 60.61 \ci{0.11} \\
ProtoNet~\cite{snell2017prototypical} & 50.11 \ci{0.31} & 64.44 \ci{0.24} &  36.44 \ci{0.17}& 46.70 \ci{0.26}\\
\hline
Triplet  & 52.53 \ci{0.66} & 63.07 \ci{0.59} & 37.24 \ci{0.37} & 49.79 \ci{0.26} \\

\hline
SimpleShot~\cite{wang2019simpleshot} & 58.99 \ci{0.29}  & 74.29 \ci{0.24} & 45.82 \ci{0.17} &  62.73 \ci{0.11}  \\
			
Shape Bias (w/ pc) - SimpleShot - $\mathcal{L}_1$ only & 59.81 \ci{0.31} & 71.61 \ci{0.26} & 47.89 \ci{0.15} & 59.48 \ci{0.11}  \\

Shape Bias (w/o pc) - SimpleShot & 60.23 \ci{0.30} & \sbest{\best{75.59 \ci{0.24}}} & 47.92 \ci{0.15} & \sbest{\best{64.88 \ci{0.11}}}   \\
Shape Bias (w/ pc) - SimpleShot &  \best{61.91 \ci{0.31}} & 75.39 \ci{0.24} & \best{49.84 \ci{0.16}} & 64.21 \ci{0.11} \\
\hline

FEAT~\cite{Ye_2020_CVPR}             & 58.30 \ci{0.29} & 71.54 \ci{0.23} & 45.41 \ci{0.16} & 60.44 \ci{0.11}\\
Shape Bias (w/o pc) - FEAT & 60.19 \ci{0.31} & 74.66 \ci{0.25}  & 48.6 \ci{0.16} & \best{64.08 \ci{0.11}}\\
Shape Bias (w pc) - FEAT & \sbest{\best{62.84 \ci{0.30}}} & \best{74.84 \ci{0.24}}  & \sbest{\best{51.49 \ci{0.15}}} & 63.80 \ci{0.11}\\

\end{tabular}}
\end{center}
\vspace{-7pt}
\caption{Results on image-only and shape-biased low-shot recognition on 
\textbf{ModelNet40-LS}. Parenthesis show confidence intervals based on 5K low shot episodes. Bold indicates best performance between a low-shot learning approach with and without shape bias; underline indicates best overall. Adding shape bias improves performance in the 1-shot learning setting and has competitive performance otherwise.}
\vspace{-5pt}

\label{tbl:modelnet-res-table}
\end{table*}

%% file: table_tex/toys_res_table.tex
\begin{table*}[h]
\renewcommand{\arraystretch}{1.15}
\begin{center}
\scalebox{0.70}{
\begin{tabular}{l|cc|cc|cc|cc}

Episode Setup $\to$  & 1-shot 5-way & 5-shot 5-way &  1-shot 10-way & 5-shot 10-way  &  1-shot 20-way & 5-shot 20-way & 1-shot 30-way & 5-shot 30-way\\
\hline
RFS~\cite{tian2020rethink} & 67.10 \ci{0.71} & 81.76 \ci{0.54} & 52.94 \ci{0.51} & 71.30 \ci{0.45} & 40.97 \ci{0.32} & 59.53 \ci{0.30} & 34.34 \ci{0.26} & 53.46 \ci{0.24}  \\
ProtoNet~\cite{snell2017prototypical} & 62.48 \ci{0.34} & 79.69 \ci{0.25} & 48.27 \ci{0.24} & 68.03 \ci{0.21} & 36.38 \ci{0.15} & 56.25 \ci{0.15} & 30.62 \ci{0.11} & 49.58 \ci{0.11}  \\

\hline

Triplet  & 63.87 \ci{0.34} & 73.95 \ci{0.62} & 48.78 \ci{0.54} & 60.44 \ci{0.48} & 36.34 \ci{0.35} & 47.28 \ci{0.32}  & 30.09 \ci{0.25} & 40.08 \ci{0.24}\\

\hline
SimpleShot~\cite{wang2019simpleshot} & 68.87 \ci{0.32} & \best{83.69 \ci{0.23}} & 55.22 \ci{0.24} & \best{73.58 \ci{0.19}} & 43.05 \ci{0.16} & \best{62.64 \ci{0.14}} & 36.78 \ci{0.12} & 56.22 \ci{0.12} \\

Shape Bias (w/o pc) - SimpleShot & 68.74 \ci{0.34} & 82.57 \ci{0.25} & 56.12 \ci{0.25} & 72.80 \ci{0.25} & 44.83 \ci{0.17} & 62.41  \ci{0.14} & 38.94 \ci{0.13} & \best{56.38 \ci{0.11}}\\
Shape Bias (w/ pc) - SimpleShot & \best{70.96 \ci{0.33}} & 81.33 \ci{0.24} & \best{58.47 \ci{0.25}} & 70.81 \ci{ 0.20}& \best{46.96 \ci{0.17}} & 60.3 \ci{0.14} & \best{40.59 \ci{0.14}} & 54.00 \ci{0.11} \\
\hline

FEAT~\cite{Ye_2020_CVPR} & 70.66 \ci{0.33} & \sbest{\best{84.13 \ci{0.23}}} & 57.15 \ci{0.24} & \sbest{\best{74.29 \ci{0.19}}} & 44.84 \ci{0.16} & \sbest{\best{63.65 \ci{0.14}}} & 38.43 \ci{0.12} & \sbest{\best{57.42 \ci{0.11}}} \\
Shape Bias (w/o pc) - FEAT & 69.21 \ci{0.32} & 82.56 \ci{0.25} & 56.76 \ci{0.24} & 72.95 \ci{0.20} & 45.15 \ci{0.16} & 62.58 \ci{0.15} & 39.24 \ci{0.12} & 56.60 \ci{0.11}\\
Shape Bias (w/ pc) - FEAT & \sbest{\best{71.58 \ci{0.34}}} & 81.45 \ci{0.25} & \sbest{\best{59.09 \ci{0.25}}} & 71.00 \ci{0.20} & \sbest{\best{47.45 \ci{0.17}}} & 59.98 \ci{0.15} & \sbest{\best{41.38 \ci{0.12}}} & 53.64 \ci{0.11} \\
\end{tabular}}
\end{center}
\vspace{-7pt}
\caption{Results on image-only and shape-biased low-shot recognition on \textbf{Toys4K}. Parenthesis show 95\% confidence intervals based on 5K low shot episodes. Bold indicates best performance for a low-shot approach with and without shape bias; underline indicates best overall. Adding shape-bias improves 1-shot performance when the number of low-shot ways is higher.}
\label{tbl:toys-res-table}
\vspace{-15pt}
\end{table*}

%% file: table_tex/shapenet_res_table.tex
\begin{table*}[h]
\renewcommand{\arraystretch}{1.15}
\begin{center}
\scalebox{0.8}{
\begin{tabular}{l|cc|cc|cc}

Episode Setup $\to$  & 1-shot 5-way & 5-shot 5-way &  1-shot 10-way & 5-shot 10-way  &  1-shot 20-way & 5-shot 20-way\\

\hline

RFS~\cite{tian2020rethink} & 65.79 \ci{0.32} & 80.51 \ci{0.23} & 52.16 \ci{0.20} & 69.92 \ci{0.10} & 40.25 \ci{0.10} & 58.44 \ci{0.08} \\
ProtoNet~\cite{snell2017prototypical} & 52.00 \ci{0.31} & 69.65 \ci{0.24} & 37.75 \ci{0.19} & 55.87 \ci{0.16} & 27.00 \ci{0.11} & 43.16 \ci{0.09}\\

\hline

Triplet  &  61.07 \ci{0.34} & 71.43 \ci{0.28} & 46.89 \ci{0.22} & 58.37 \ci{0.18} & 35.09 \ci{0.12} & 46.20 \ci{0.08}  \\

\hline

SimpleShot~\cite{wang2019simpleshot} & 66.73 \ci{0.32} & 80.93 \ci{0.22} & 53.37 \ci{0.21} & 70.32 \ci{0.16} & 41.09 \ci{0.12} & 59.09 \ci{0.08}\\
Shape Bias (w/o pc) - SimpleShot & 67.5 \ci{0.34} & \best{81.30 \ci{0.23}} & 54.99 \ci{0.23} & \best{71.24 \ci{0.17}} & 43.60 \ci{0.13} & \best{61.03 \ci{0.08}}\\
 Shape Bias (w/ pc) - SimpleShot & \best{69.72 \ci{0.32}} & 80.93 \ci{0.24} & \best{57.49 \ci{0.21}} & 70.75 \ci{0.16} & \best{46.24 \ci{0.12}} & 60.21 \ci{0.08} \\

\hline

FEAT~\cite{Ye_2020_CVPR} & 67.81  \ci{0.32} & 80.25 \ci{0.23} & 54.35 \ci{0.22} & 70.18 \ci{0.16} & 42.12 \ci{0.12} & 59.01 \ci{0.08} \\
Shape Bias (w/o pc)- FEAT & 67.78 \ci{0.32} & \sbest{\best{81.45 \ci{0.22}}} & 55.69 \ci{0.22} & \sbest{\best{71.74 \ci{0.16}}} & 44.44 \ci{0.13} & \sbest{\best{61.46 \ci{0.08}}} \\
Shape Bias (w/ pc) - FEAT & \sbest{\best{70.24 \ci{0.32}}} & 80.95 \ci{0.22} & \sbest{\best{58.45 \ci{0.22}}} & 70.95 \ci{0.16} & \sbest{\best{47.03 \ci{0.13}}} & 60.43 \ci{0.08} \\
\end{tabular}}
\end{center}
\vspace{-7pt}
\caption{Results on image-only and shape-biased low-shot recognition on ShapeNet55-LS. Parenthesis show confidence intervals based on 5K low shot episodes. Bold indicates best performance between a low-shot approach with and without shape bias and underline indicates best overall. Adding shape bias leads to consistent improvement for both FEAT and SimpleShot.}
\label{tbl:shapenet-res-table}
\vspace{-15pt}
\end{table*}

%% file: figure_tex/pairwise_loss_analysis.tex
\begin{figure}[t]
\begin{center}
\includegraphics[width=\linewidth]{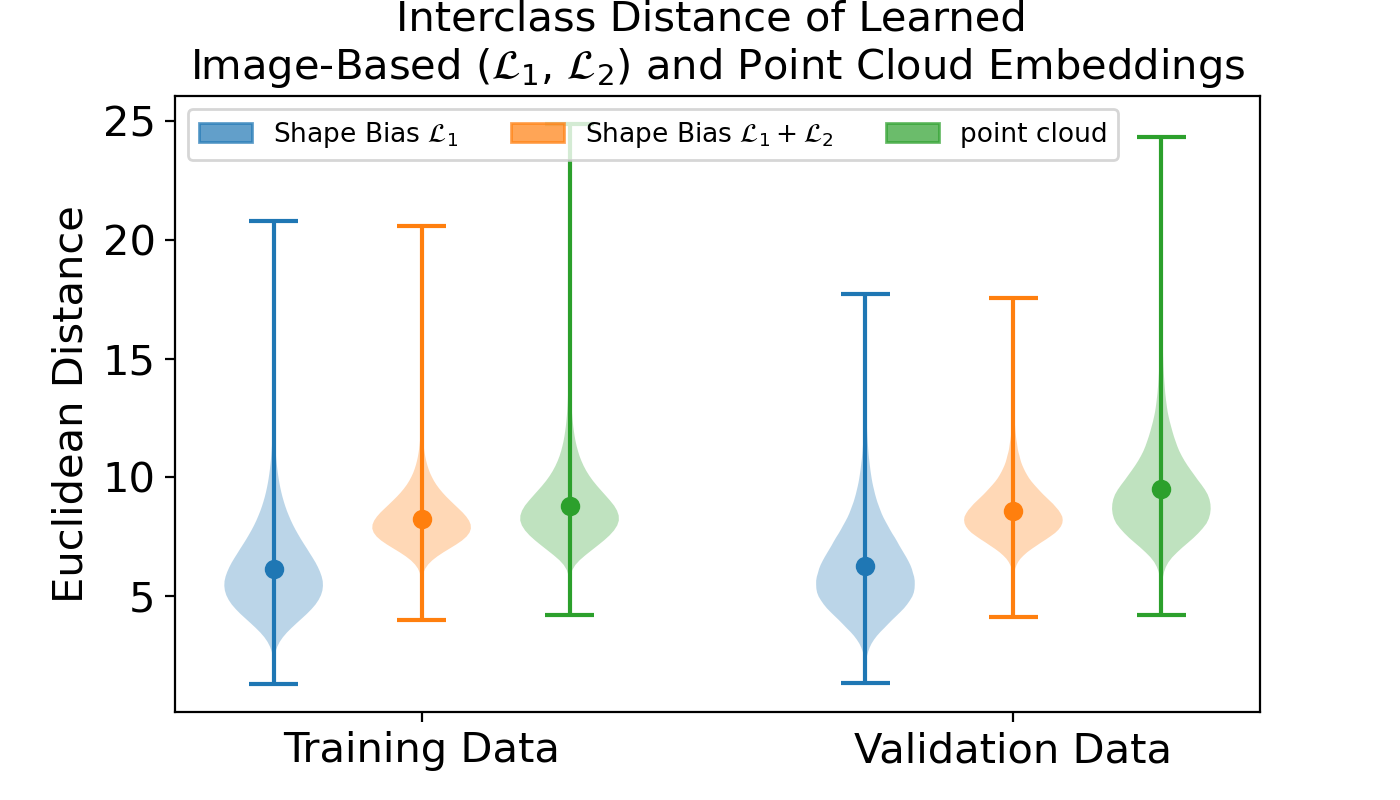}
\end{center}
\vspace{-12pt}
\caption{Examining the distribution of interclass distances in the mappings learned by minimizing either $\mathcal{L}_1$ or $\mathcal{L}_1 + \mathcal{L}_2 $ relative to the reference point cloud embedding shows that adding $\mathcal{L}_2$ in a better approximation of the shape embedding space on both novel categories and categories seen during training.}
\label{fig:pairwise-analysis}
\vspace{-18pt}
\end{figure}

%% file: figure_tex/accuracy_correlation_fig.tex
\begin{figure}[t]
\begin{center}
\includegraphics[width=\linewidth]{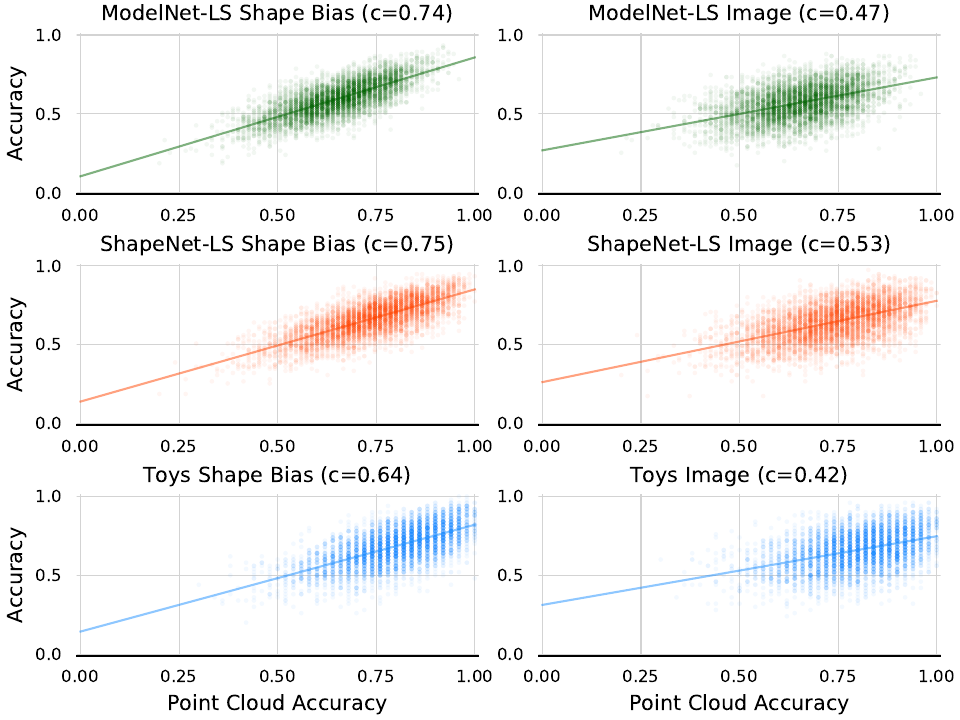}
\end{center}
\vspace{-12pt}
\caption{Accuracy of point cloud SimpleShot model vs. shape bias and image SimpleShot models over the same 5K episodes shows higher correlation between the point cloud model and shape bias model, indicating that the shape-biased model classifies more similarly to the point cloud model than the image only model.}
\label{fig:correlation-analysis}
\vspace{-16pt}
\end{figure}

%% file: table_tex/percent_misclassified.tex
\begin{table}[ht!]
\centering
\begin{tabular}{l|c|c|c}
& ModelNet & ShapeNet & Toys \\
\hline
5-way  & 38.73\% & 44.81\% & 57.59\% \\
10-way & 30.88\% & 38.17\% & 50.53\% \\
\end{tabular}
\vspace{5pt}
\caption{Percent of queries misclassified by shape-biased SimpleShot but not misclassified by point cloud model (over 5K episodes). This indicates there is significant room for improvement by learning better maps from images to shape embeddings.}
\label{tbl:misclassification-analysis}
\vspace{-10pt}
\end{table}

%% file: sections/4_conclusion.tex
\vspace{-5pt}
\section{Discussion and Conclusion}

This paper takes the first step in investigating the utility of shape bias for low-shot object categorization. Through extensive empirical analysis of our novel approach for adding shape bias to image-only low-shot learning algorithms, we demonstrate improved generalization. We also introduce Toys4K, a diverse and challenging dataset for object learning with the largest number of categories available to date. While dependence of our findings on synthetic object data limits our ability to draw conclusions about shape bias under more general conditions, it is essential since it is currently the only feasible way to obtain matched 2D and 3D data at a large enough scale. Moreover, synthetic data has been widely adopted for other vision tasks~\cite{deitke2020robothor, dosovitskiy2017carla, johnson2017clevr}.

Progress in few-shot learning is crucial in order to overcome the need for large amounts of labeled training data. This work constitutes a step in a new direction: the exploitation of the natural biases of the visual world, such as object shape, in the design of few-shot architectures. Building on this approach by exploiting other sources of bias is a logical and exciting direction for future work.

%% file: sections/appendix.tex
\begin{center}
{\scshape\LARGE Appendix \par}
\end{center}

This supplementary material document is structured as follows: In Section~\ref{sec:datasets} we provide further detail about the training data used in the paper; In section~\ref{sec:baselines} we provide details on the baselines used in the paper, their implementation details and the hyperparameters used for training; In Section~\ref{sec:ptcld-embedding} we provide empirical evidence about our choice of point-cloud encoding architecture; In Section~\ref{sec:shape-bias} we provide further details about the training procedure of the shape-biased image embeddings used in the paper.

\section{Further Dataset Details}
\label{sec:datasets}
\begin{figure*}[h]
\begin{center}
\includegraphics[width=\linewidth]{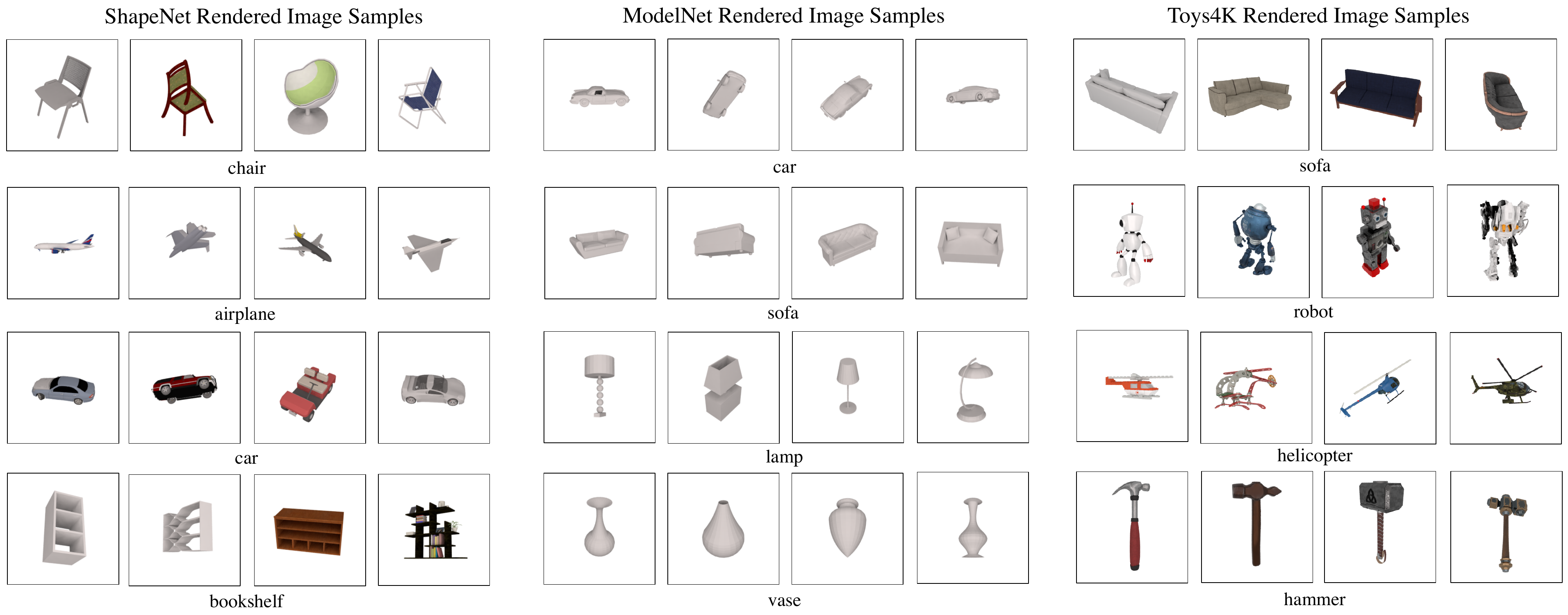}
\end{center}
\caption{Rendered image samples from multiple categories of ShapeNet, ModelNet and Toys4K. Note the high image quality as a result of using ray-tracing based rendering.}
\label{fig:render_samples}
\end{figure*}

In this section we provide details on the composition of the datasets used in the main paper. We provide example images used to illustrate the data used for training in Figure~\ref{fig:render_samples}. As a result of using a ray tracing-based renderer Cycles~\citeNew{blender1}, the synthetic image data used for training has high realism. For all algorithms we use $224 \times 224$ RGB images as input. For point cloud-based learning we use the 3D $(x,y,z)$ coordinates 1024 randomly sampled points as input. For images we use standard geometric data augmentations e.g. flipping, cropping, slight translation and rotation, as well as color jittering, since we found these result in improved validation performance. For point clouds we use the same augmentation procedures as in~\citeNew{pointnet2-repo}, which include translation, jittering and dropout.
\subsection{Toys4K}
We provide further details on the composition of our new Toys4K dataset in Table~\ref{tbl:dataset-composition}. The 40 train, 10 validation, and 55 test classes split is shown in Table~\ref{tbl:toys-split}. When performing validation and testing on Toys4K, we generate low-shot episodes consisting of up to 5 shots and 10 queries.
\subsection{ModelNet40-LS}
The 20 train, 10 validation, 10 test classes split for ModelNet40-LS is shown in Table~\ref{tbl:modelnet-split}. When performing validation and testing on ModelNet40-LS, we generate low-shot episodes consisting of up to 5 shots and 15 queries.
\subsection{ShapeNet55-LS}
The 25 train, 10 validation, 20 test classes split for ShapeNet55-LS is shown in Table~\ref{tbl:shapenet-split}. When performing validation and testing on ShapeNet55-LS, we generate low-shot episodes consisting of up to 5 shots and 15 queries.

\begin{table*}[h]
\begin{center}
\scalebox{0.95}{
\begin{tabular}{lr|lr|lr|lr|lr|lr|lr}
chair       & 210 & tree         & 57 & knife         & 45 & piano    & 39 & shark   & 30 & panda      & 24 & submarine   & 18 \\
bottle      & 111 & candy        & 56 & trashcan      & 44 & boat     & 38 & stove   & 29 & orange     & 24 & helmet      & 17 \\
robot       & 105 & guitar       & 55 & ball          & 44 & bread    & 38 & bowl    & 28 & mushroom   & 23 & bicycle     & 16 \\
dog         & 103 & apple        & 54 & frog          & 43 & fish     & 37 & car     & 28 & phone      & 23 & lion        & 16 \\
mug         & 97  & flower       & 54 & ice cream     & 43 & horse    & 36 & cookie  & 28 & train      & 22 & motorcycle  & 16 \\
hammer      & 94  & ladder       & 53 & dragon        & 43 & spade    & 36 & cupcake & 28 & tv         & 21 & hamburger   & 16 \\
cat         & 79  & penguin      & 51 & pan           & 42 & banana   & 35 & bunny   & 27 & toaster    & 21 & grapes      & 16 \\
dinosaur    & 76  & keyboard     & 51 & battery cell  & 41 & airplane & 35 & drum    & 26 & helicopter & 20 & tractor     & 16 \\
deer/moose  & 65  & pencil       & 50 & whale         & 41 & donut    & 34 & pizza   & 26 & lizard     & 20 & monkey      & 16 \\
fox         & 64  & plate        & 50 & shoe          & 40 & truck    & 34 & mouse   & 25 & saw        & 19 & pc mouse    & 15 \\
hat         & 64  & key          & 49 & laptop        & 40 & coin     & 33 & chicken & 25 & marker     & 19 & light bulb  & 15 \\
sofa        & 63  & chess piece  & 49 & pig           & 40 & snake    & 32 & sink    & 25 & microwave  & 18 & closet      & 15 \\
glass       & 63  & cake         & 48 & sheep         & 39 & fridge   & 32 & cow     & 25 & bus        & 18 & fries       & 15 \\
cup         & 60  & screwdriver  & 46 & crab          & 38 & octopus  & 31 & dolphin & 25 & pear       & 18 & sandwich    & 15 \\
monitor     & 57  & elephant     & 46 & radio         & 38 & fan      & 31 & violin  & 25 & butterfly  & 18 & giraffe     & 15 \\
\end{tabular}}
\end{center}
\caption{The category composition of the Toys4K dataset. }
\label{tbl:dataset-composition}
\end{table*}

\section{Further Low-Shot Analysis}
\label{sec:further-analysis}

In this section we provide further anaylsis of the low-shot performance by presenting confusion matrices and classification performance in invidividual low-shot episodes.

\subsection{Confusion Matrices}

\begin{figure*}[t!]
\begin{center}
\includegraphics[width=\linewidth]{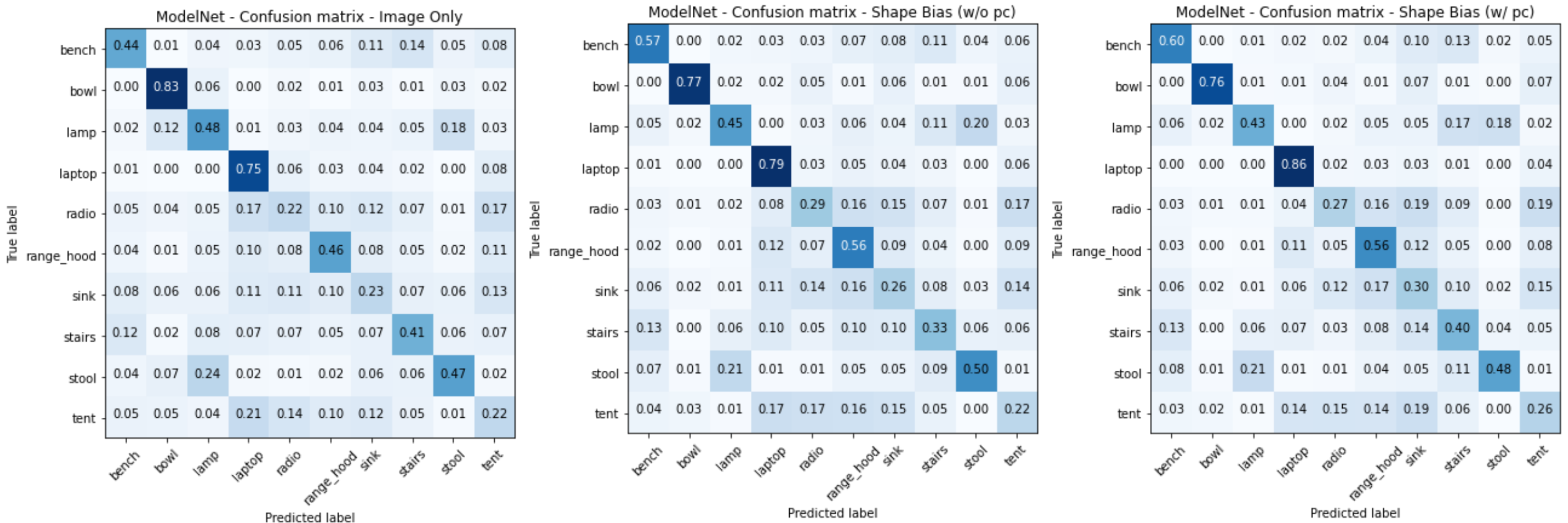}
\end{center}
\caption{Confusion matrices over 5K low-shot episodes of SimpleShot for Image Only, Shape-Biased without access to point clouds (w/o pc) at test time and Shape-Biased with (w/ pc) access to point clouds at test time on the ModelNet-LS dataset. Even without access to point clouds (w/o pc) for building class prototypes, the shape-biased image embedding leads to improvements. Adding point cloud support information (w/ pc) improves performance further. See Table~\ref{tbl:modelnet-res-table} for aggregate results.}
\label{fig:conf_mat_modelnet}
\end{figure*}

\begin{figure*}[t!]
\begin{center}
\includegraphics[width=\linewidth]{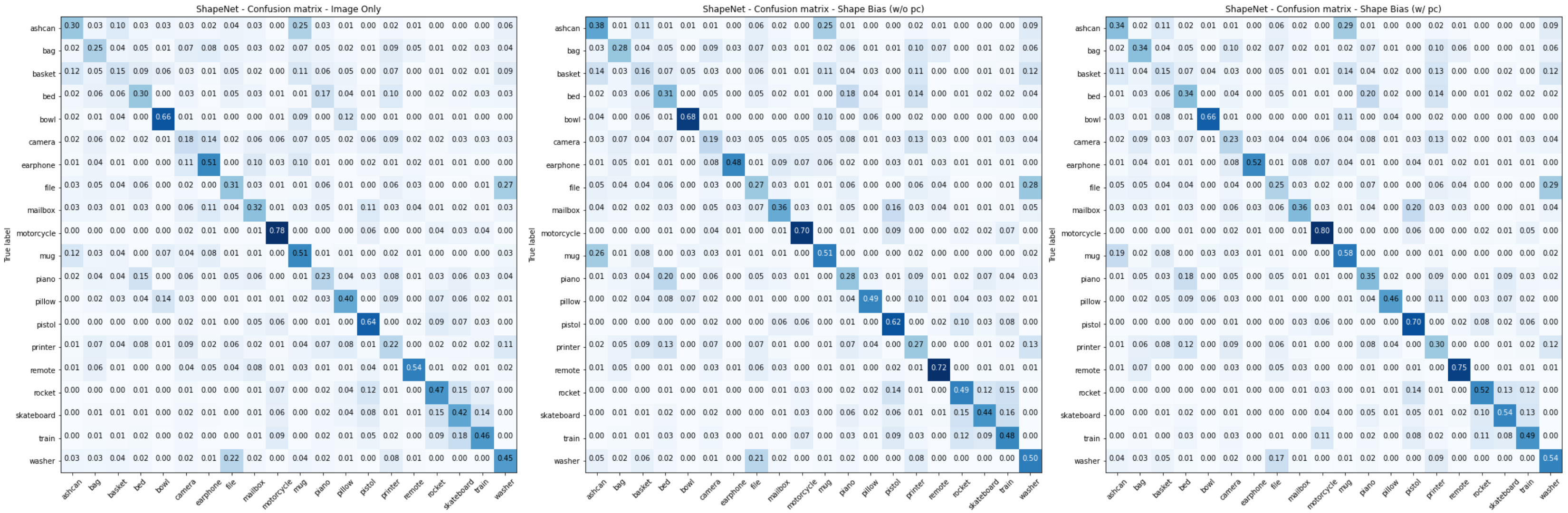}
\end{center}
\caption{Confusion matrices over 5K low-shot episodes of SimpleShot for Image Only, Shape-Biased without access to point clouds (w/o pc) at test time and Shape-Biased with (w/ pc) access to point clouds at test time on the ShapeNet-LS dataset. As in ModelNet40-LS, without access to point clouds (w/o pc) for building class prototypes, the shape-biased image embedding leads to improvements. Adding point cloud support information (w/ pc) improves performance further. See Table~\ref{tbl:shapenet-res-table} for aggregate results. Best viewed with zoom.}
\label{fig:conf_mat_shapenet}
\end{figure*}

Please refer to Figure~\ref{fig:conf_mat_modelnet} and Figure~\ref{fig:conf_mat_shapenet} for low-shot confusion matrices on ModelNet40-LS and ShapeNet55-LS. The confusion matrices are obtained by evaluation 5K low-shot episodes for each dataset (10-way for ModelNet40-LS and 20-way for ShapeNet55-LS), and counting how each sample was classified. The confusion matrices reflect the results presented in Section~\ref{sec:experiments} in the main text that adding shape bias improves overall low-shot classification performance.

\subsection{Per-episode Analysis}
\begin{figure*}[t!]
\begin{center}
\includegraphics[width=\linewidth]{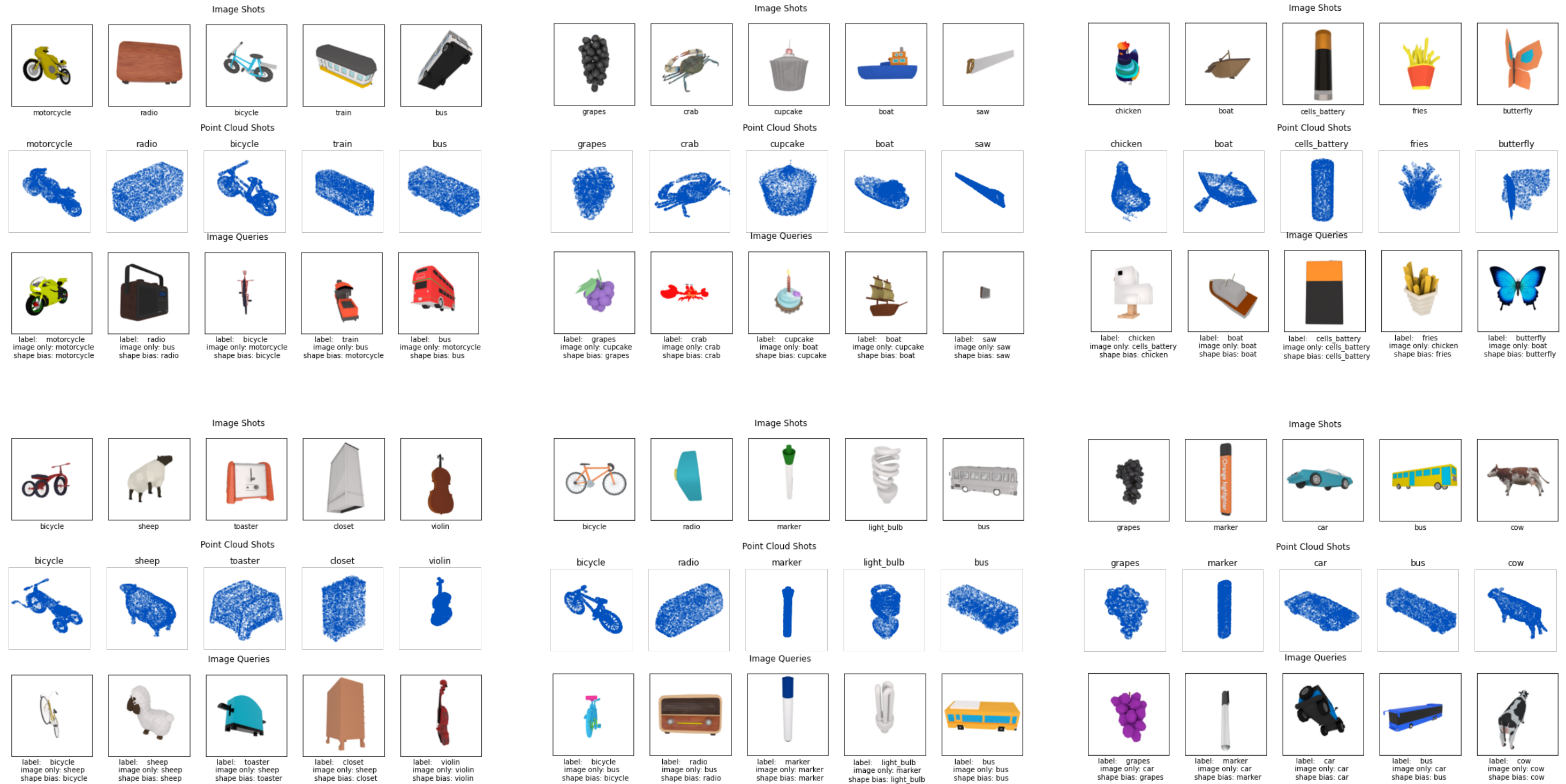}
\end{center}
\caption{Six low-shot episodes for 5 ways, 1 shot and 1 query on Toys4K for shape-biased SimpleShot. We visually display the composition of the image and point cloud shots and the image queries, as well as the models' predictions, illustrating cases where shape bias allows for improved performance. Best viewed with zoom.}
\label{fig:episode_analysis}
\end{figure*}

We provide a per-episode analysis of low-shot classification in Figure~\ref{fig:episode_analysis} to show qualitative evidence of low-shot learning with shape bias. We see that there are cases in which even though there are no view ambiguities, the image-only model misclassifies whereas the shape-biased model correctly classifies (e.g. in the lower left episode, confusing bicycle for sheep).

\section{Baseline Algorithm Details}
\label{sec:baselines}
All algorithms in this paper are implemented using PyTorch~\citeNew{paszke2019pytorch}. In this section we provide further detail about the baseline implementations and hyperparameters used for training.
\subsection{SimpleShot}
The implementation in our codebase for SimpleShot~\citeNew{wang2019simpleshot1} is based on the code release by the authors in \citeNew{simpleshot-repo}. The authors report a 1-shot 5-way accuracy of $49.69(0.19)$ and a 
5-shot 5-way accuracy of $66.92(0.17)$ on miniImageNet~\citeNew{vinyals2016matching1} with the Conv4 architecture. The reimplementation of SimpleShot in our codebase with the same dataset and architecture results in 1-shot 5-way accuracy of $50.60(0.34)$ and a 
5-shot 5-way accuracy of $68.06(0.23)$. 

In all our experiments we train SimpleShot with SGD with an initial learning rate of $0.01$ and a learning rate decay of $0.1$ at epochs 300 and 360, out of a total of 400 epochs. SimpleShot employs three different feature normalization strategies, no normalization, $L_2$ normalization and $L_2$ normalization and training set mean subtraction. In experiments with SimpleShot we report the result of the best of these three normalization strategies.

\subsection{RFS}
The implementation in our codebase for RFS~\citeNew{tian2020rethink1} is based on the code release by the authors in~\citeNew{RFS-repo}. The original codebase obtains a 1-shot 5-way accuracy of $53.73 (0.81)$ on miniImageNet~\citeNew{vinyals2016matching1} with the Conv4 architecture. The reimplementation of RFS in our codebase with the same dataset and architecture results in 1-shot 5-way accuracy of $ 54.59(0.86)$. 
RFS requires training an embedding on the training dataset using cross-entropy. We train this embedding space with SGD using a learning rate of 0.001, momentum of 0.9 and $L_2$ weight penalty weight parameter of 0.0005. For each low-shot episode we train a logistic regression classifier using Scikit-learn\citeNew{scikit-learn}, as in the original RFS.
\subsection{FEAT}
The implementation for FEAT is based on the code release by the authors in \citeNew{FEAT-repo}. The original codebase obtains a 1-shot 5-way accuracy of $54.85 (0.20)$ and 5-shot 5-way accuracy of 
$71.61$ on miniImageNet~\citeNew{vinyals2016matching1} with the Conv4 architecture. The reimplementation of FEAT in our codebase with the same dataset and architecture results in 1-shot 5-way accuracy of $54.85 (0.20)$ 5-shot 5-way accuracy of $71.45 (0.73)$. We train FEAT with the default hyperparameters recommended by the authors, training separate models for 5-way and 10-way classification, and separate models for 1-shot and 5-shot, as recommended by the authors. For the shape biased FEAT we do not use learning rate scheduling and momentum, since they have a negative effect on performance for shape-biased training. Removing them for image-only training does not affect performance.

\subsection{Protoypical Networks}
The implementation in our codebase for Prototypical Networks is based on the code release by the SimpleShot authors in \citeNew{simpleshot-repo}. In \citeNew{chen2018closer} the authors report that their reimplementation obtains a 5-shot 5-way accuracy of $66.68 (0.68)$ on miniImageNet~\citeNew{vinyals2016matching1} with the Conv4 architecture. The reimplementation of Prototypical Networks in our codebase with the same dataset and architecture results in 5-shot 5-way accuracy of $ 66.94 (0.71)$. We train separate Prototypical Networks models for 5-shot classification and 1-shot classification. As recommended by the authors of the original paper, we perform 20-way training. We use the Adam~\citeNew{kingma2014adam} optimizer, 400 low-shot iterations per epoch, 200 epochs total, and a learning rate of 0.0001 $L_2$ and weight penalty weight parameter of 0.00001. We perform a learning rate decay of 0.5 every 20 epochs.

\subsection{Triplet Model}
We implement a joint triplet model using both point cloud (DGCNN~\citeNew{wang2019dynamic1} and image (ResNet18~\citeNew{he2016deep1}) encoders, which can use both image and shape information during training. Let $f_i$ denote the image encoder, $f_p$ denote the point cloud encoder, and $\phi_p^k$ and $\phi_i^k$ denote the point cloud and image encodings of object instance $k$ respectively. We learn a joint image/shape embedding by minimizing a standard triplet loss
\begin{align*}
    \mathcal{L}(\phi_i^k, \phi_p^k, \phi_p^l) &= \max\big\{\mathbf{d}(\phi_i^k, \phi_p^k) - \mathbf{d}(\phi_p^k, \phi_p^l)+\text{margin}, 0\big\}\\
    \mathbf{d}(x,y) &= ||x-y||_2
\end{align*}
where the anchor is an image embedding of instance $k$, $\phi_i^k$, the positive sample is a point cloud encoding of the same object instance, $\phi_i^k$, and the negative sample is a $\phi_p^l$ is a point cloud embedding of a different object instance $l$. We perform $L_2$ normalization of the embeddings prior to computing the loss. Note that it is possible to build (anchor, positive, negative) pairs using category information, but we empirically found that this leads to worse performance.

We train the triplet model using the Adam~\citeNew{kingma2014adam} optimizer with a learning rate of 0.0001,  $L_2$ weight penalty weight parameter of 0.0001, and margin of 0.1. We use a batch size of 72 and train for 600 epochs, each epoch consisting of 20K random samples.

\section{Learning a Point Cloud Shape Embedding}
\label{sec:ptcld-embedding}
In this section we describe the algorithm for learning a point-cloud based embedding space, and present an empirical study for our point cloud architecture choice.
\subsection{Algorithm}
The algorithm we use to train a point-cloud embedding space is based on SimpleShot~\citeNew{wang2019simpleshot1} and is described with pseudocode in Algorithm~\ref{alg:shape_embedding}. Note that the routine AccAccumulator is used denotes a function to collect the validation accuracy of each low-shot episode and compute summary statistics. The \textsc{NNClassify} routine takes support features and labels, and classifies each test query feature based on a nearest neighbors rule using cosine similarity. The point-cloud embedding model is trained using SGD with a learning rate of 0.01, batch size of 129, and $L_2$ weight penalty weight parameter of 0.0001. We perform learning rate decay by 0.1 at epochs 300 and 360. In all models we use features from the output of the pooling layer in the architecture.
\subsection{Architecture Study}
We perform an empirical study on the point cloud architectures to determine which is capable of the best low-shot generalization performance. Our PointNet~\citeNew{qi2017pointnet1} implementation is based on  \citeNew{pointnet-repo}, our PointNet++~\citeNew{qi2017pointnet++1} is based on~\citeNew{pointnet2-repo} and our DGCNN~\citeNew{wang2019dynamic1} implementation is based on~\citeNew{dgcnn-repo}. We use a DGCNN architecture with a reduced embedding dimension (size after the pooling operation) of 512 rather than the original 1024, to match the dimensionality of the ResNet18 embeddings. We find no decrease in performance by this reduction. We present the results of this study on ModelNet in Table~\ref{tbl:arch-analysis}. The DGCNN~\citeNew{wang2019dynamic1} architecture outperforms other point cloud architectures at low-shot generalization to novel categories. We find that randomly rotating the input point cloud about the origin during training (random rotation about all axes of rotation, indicated by SO3 in the table) results in a performance improvement. We use this SO3 strategy for all shape-embedding space learning experiments.

\begin{table}[hbt!]
\begin{center}
\scalebox{0.95}{
\begin{tabular}{l|c}
Architecture & 1-shot 5-way accuracy \\
\hline
PointNet~\citeNew{qi2017pointnet1}  & 66.13 \\
PointNet++~\citeNew{qi2017pointnet++1} & 67.49 \\
DGCNN~\citeNew{wang2019dynamic1} & 75.2 \\
DGCNN (SO3) & 77.5 \\
\end{tabular}}
\end{center}
\caption{Empirical study for choosing the best point cloud architecture. Reported is 1-shot 5-way classification accuracy on the ModelNet40-LS validation set. We find that DGCNN performs the best, and that randomly rotating each input point cloud during training (indicated with SO3) results in a improvement in low-shot generalization performance as well.}
\label{tbl:arch-analysis}
\end{table}

\section{Details for Learning a Shape Biased Image Embedding}
\label{sec:shape-bias}
The algorithm we use to train a shape-biased image embedding is described with pseudocode in Algorithm~\ref{alg:shape_biased_embedding}. We use the Adam optimizer with a batch size of 256, an initial learning rate of 0.001 and a $L_2$ weight penalty weight parameter of 0.0001. The model is trained for 400 epochs, with a learning rate decay of 0.1 at epochs 300 and 360.

\subsection{SimpleShot with Shape Bias}
The SimpleShot~\citeNew{wang2019simpleshot1} approach does not require any learning (parameter updates) during the low-shot phase. Classification is done using nearest centroid classification in the embedding space. The image embedding function $f_i$ is trained as described in Algorithm~\ref{alg:shape_biased_embedding}, and low-shot testing is done following the same procedure as described in L8-16 in Algorithm~\ref{alg:shape_embedding} but using nearest centroid rather than nearest neighbor classification.

\subsection{FEAT with Shape Bias}
The algorithm we use to train a shape-biased FEAT~\citeNew{Ye_2020_CVPR1} architecture is described in Algorithm~\ref{alg:shape_biased_FEAT}. Note that the $f_i$ used in this algorithm is being fine tuned from a mapping already trained with Algorithm~\ref{alg:shape_biased_embedding} while the FEAT set-to-set function $\mathbf{E}$ is trained from scratch. For this experiment we use the default hyperparameters recommended by the FEAT authors. Low shot testing is done following the same procedure as described in L13-22 in Algorithm~\ref{alg:shape_biased_FEAT} but using the test set. The procedure we refer to as \textsc{FEATClassify} is described in Eq. 4 on pg. 4 of the FEAT paper~\citeNew{Ye_2020_CVPR1}. In the pseudocode \textsc{FEATClassify} performs classification and directly outputs the per-episode classification accuracy.

\clearpage
\begin{algorithm}[t!]
\label{alg:shape_embedding}
\caption{Training Shape Embedding $f_p$}
\SetAlgoLined
\KwIn{Randomly initialized point-cloud classifier architecture $f_p$ with embedding function $f_p^{E}$\newline
      Total number of epochs $N_e$ \newline
      Total number of mini-batches per epoch $N_b$\newline
      Total number of low-shot iterations for validation $N_{it}$}
\KwData{(point cloud, label) pair datasets $\mathcal{D}^{\text{train}}$,  $\mathcal{D}^{\text{val}}$}
\KwDef{$\ell:$ cross-entropy loss }
\ForEach{\text{epoch in $1,2,\dots,N_e$} }{
  \ForEach{mini-batch $(\bm{o}_p, \bm{y})\sim \mathcal{D}^{\text{train}}$ of $N_b$}{
  Predict $\hat{\bm{y}} = f_p(\bm{o}_p)$\\
  Compute $\ell(\bm{y}, \hat{\bm{y}})$\\
  Compute $\nabla \ell$ with respect to $f_p$\\
  Update $f_p$ with SGD\\
  }
  A = \textsc{AccAccumulator}\\
  \ForEach{validation episode in $1,2,\dots,N_e$}{
    Sample $5$-way $1$-shot $(\bm{o}_p^\text{train}, \bm{y}^\text{train}, \bm{o}_p^\text{test}) \sim \mathcal{D}^{\text{val}}$\\
    Predict $\bm{\phi}_p^\text{train} = f_p^E(\bm{o}_p^\text{train})$\\
    Predict $\bm{\phi}_p^\text{test} = f_p^E(\bm{o}_p^\text{test})$\\
    acc = \textsc{NNClassify}($\bm{\phi}_p^\text{train}, \bm{y}^\text{train}, \bm{\phi}_p^\text{test}$)\\
    A(acc)
  }
  val accuracy = A.average()\\
  \If{val accuracy $>$ best accuracy}{
    best acuracy $\leftarrow$ val accuracy\\
    $f_p^\text{best} \leftarrow f_p$
  }
 }
\KwResult{Trained $f_p^\text{best}$}
\end{algorithm}

\begin{algorithm}[t!]
\label{alg:shape_biased_embedding}
\caption{Training Shape-Biased Image \\ Embedding Function $f_i$}
\SetAlgoLined
\KwIn{Randomly initialized image embedding architecture $f_i$ \newline
      Point-cloud embedding function $f_p$ (\ref{alg:shape_embedding})\newline
      Total number of epochs $N_e$ \newline
      Total number of mini-batches per epoch $N_b$\newline
      Total number of low-shot iterations for validation $N_{it}$}
\KwData{(image, point cloud, label) pair datasets $\mathcal{D}^{\text{train}}$,  $\mathcal{D}^{\text{val}}$}
\KwDef{$\mathcal{L} = \mathcal{L}_1 + \mathcal{L}_2 $ (see main text for def.)}
\ForEach{\text{epoch in $1,2,\dots,N_e$} }{
  \ForEach{mini-batch $(\bm{o}_i, \bm{o}_p, \bm{y})\sim \mathcal{D}^{\text{train}}$ of $N_b$}{
  Predict shape embedding $\bm{\phi}_{p} = f_p(\bm{o}_p)$\\
  Predict image embedding $\bm{\phi}_{i} = f_i(\bm{o}_i)$\\
  Compute $\mathcal{L}$ using $\bm{\phi}_{p}$ and $\bm{\phi}_{i}$\\
  Compute $\nabla \mathcal{L}$ with respect to $f_i$\\
  Update $f_i$ with Adam\\
  }
  A = \textsc{AccAccumulator}\\
  \ForEach{validation episode in $1,2,\dots,N_e$}{
    Sample $5$-way $1$-shot $(\bm{o}_i^\text{train}, \bm{o}_p^\text{train}, \bm{y}^\text{train}, \bm{o}_i^\text{test}) \sim \mathcal{D}^{\text{val}}$\\
    Predict $\bm{\phi}_p^\text{train} = f_p(\bm{o}_p^\text{train})$ \\
    Predict $\bm{\phi}_i^\text{train} = f_i(\bm{o}_i^\text{train})$\\
    Predict $\bm{\phi}_i^\text{test} = f_i(\bm{o}_i^\text{test})$\\
    $\bm{\phi}^\text{train} \leftarrow$ \textsc{Average}$(\bm{\phi}_p^\text{train},\bm{\phi}_i^\text{train})$\\
    acc = \textsc{NNClassify}($\bm{\phi}^\text{train}, \bm{y}^\text{train}, \bm{\phi}_i^\text{test}$)\\
    A(acc)
  }
  val accuracy = A.average()\\
  \If{val accuracy $>$ best accuracy}{
    best acuracy $\leftarrow$ val accuracy\\
    $f_i^\text{best} \leftarrow f_i$
  }
 }
\KwResult{Trained $f_i^\text{best}$}
\end{algorithm}

\begin{algorithm}[t!]
\label{alg:shape_biased_FEAT}
\caption{Training FEAT with Shape Bias}
\SetAlgoLined
\KwIn{Shape-biased image encoder $f_i$ (\ref{alg:shape_biased_embedding})\newline
      Point-cloud embedding function $f_p$ (\ref{alg:shape_embedding})\newline
      Randomly initialized FEAT~\citeNew{Ye_2020_CVPR1} set-to-set function $\mathbf{E}$--see p3 in~\citeNew{Ye_2020_CVPR1}.\newline
      Total number of epochs $N_e$ \newline
      Total number of low-shot iterations per training epoch $N_{it}$\newline
      Total number of low-shot iterations for validation $N_{v-it}$}
\KwData{(image, point cloud, label) pair datasets $\mathcal{D}^{\text{train}}$,  $\mathcal{D}^{\text{val}}$}
\KwDef{$\mathcal{L}_{\text{FEAT}}$ -- Eq. 7 in~\citeNew{Ye_2020_CVPR1}}
\ForEach{\text{epoch in $1,2,\dots,N_e$} }{
  \ForEach{training episode in of $1,2,\dots,N_{it}$}{
  Sample $m$-way $n$-shot $(\bm{o}_i^\text{train}, \bm{o}_p^\text{train}, \bm{o}_p^\text{query}, \bm{y}^\text{train}, \bm{y}^\text{query}) \sim \mathcal{D}^{\text{train}}$\\
  Predict ptcld. support $\bm{\phi}_p^\text{train} = f_p(\bm{o}_p^\text{train})$ \\
  Predict image support $\bm{\phi}_i^\text{train} = f_i(\bm{o}_i^\text{train})$\\
  Predict image queries $\bm{\phi}_i^\text{query} = f_i(\bm{o}_i^\text{query})$\\
  $\bm{\phi}^\text{train} \leftarrow$ \textsc{Average}$(\bm{\phi}_p^\text{train},\bm{\phi}_i^\text{train})$\\
  $\hat{\bm{\phi}}^\text{train}, \hat{\bm{\phi}}_i^\text{query} \leftarrow \mathbf{E}(\bm{\phi}^\text{train}, \bm{\phi}_i^\text{query})$\\
  Compute $\mathcal{L}$ using $\hat{\bm{\phi}}^\text{train}, \hat{\bm{\phi}}_i^\text{query}$ and $\bm{y}^\text{train}, \bm{y}^\text{query}$\\
  Compute $\nabla \mathcal{L}$ with respect to $f_i$ and $\mathbf{E}$\\
  Update $f_i, \mathbf{E}$ with SGD\\
  }
  A = \textsc{AccAccumulator}\\
  \ForEach{validation episode in $1,2,\dots,N_{v-it}$}{
    Sample $5$-way $1$-shot $(\bm{o}_i^\text{train}, \bm{o}_p^\text{train}, \bm{y}^\text{train}, \bm{o}_i^\text{test}) \sim \mathcal{D}^{\text{val}}$\\
    Predict ptcld. support $\bm{\phi}_p^\text{train} = f_p(\mathbf{o}_p^\text{train})$ \\
    Predict image support $\bm{\phi}_i^\text{train} = f_i(\mathbf{o}_i^\text{train})$\\
    Predict image queries $\bm{\phi}_i^\text{test} = f_i(\mathbf{o}_i^\text{test})$\\
    $\bm{\phi}^\text{train} \leftarrow$ \textsc{Average}$(\bm{\phi}_p^\text{train},\bm{\phi}_i^\text{train})$\\
    acc = \textsc{FEATClassify}($\bm{\phi}^\text{train}, \bm{y}^\text{train}, \bm{\phi}_i^\text{test}$)\\
    A(acc)
  }
  val accuracy = A.average()\\
  \If{val accuracy $>$ best accuracy}{
    best acuracy $\leftarrow$ val accuracy\\
    $f_i^\text{best} \leftarrow f_i$\\
    $\mathbf{E}^\text{best} \leftarrow \mathbf{E}$
  }
 }
\KwResult{Trained $f_i^\text{best}, \mathbf{E}^\text{best}$}
\end{algorithm}

\clearpage

\begin{table}[t!]
\begin{center}
\setlength{\tabcolsep}{3.5pt}
\scalebox{0.80}{
\begin{tabular}{|lr|lr|lr|}
\hline
Training & \# samples & Validation & \# samples  & Testing & \# samples \\
\hline
\hline
vessel      & 873 & train       & 389 & mug         & 214  \\
car         & 530 & bed         & 233 & tower       & 133  \\
sofa        & 500 & stove       & 218 & motorcycle  & 337  \\
lamp        & 500 & bowl        & 186 & cap         & 56   \\
cellular    & 500 & pillow      & 96  & pistol      & 307  \\
faucet      & 500 & mailbox     & 94  & earphone    & 73   \\
pot         & 500 & rocket      & 85  & skateboard  & 152  \\
guitar      & 500 & birdhouse   & 73  & camera      & 113  \\
airplane    & 500 & microphone  & 67  & piano       & 239  \\
bus         & 500 & keyboard    & 65  & printer     & 166  \\
chair       & 500 &             &     & bag         & 83   \\
rifle       & 500 &             &     & trashcan    & 343  \\
cabinet     & 500 &             &     & file        & 298  \\
bench       & 499 &             &     & dishwasher  & 93   \\
bathtub     & 499 &             &     & microwave   & 152  \\
telephone   & 499 &             &     & washer      & 169  \\
jar         & 499 &             &     & remote      & 66   \\
bottle      & 498 &             &     & helmet      & 162  \\
display     & 496 &             &     & basket      & 113  \\
clock       & 496 &             &     & can         & 108  \\
loudspeaker & 496 &             &     &             &      \\
table       & 495 &             &     &             &      \\
laptop      & 460 &             &     &             &      \\
bookshelf   & 452 &             &     &             &      \\
knife       & 423 &             &     &             &      \\
\hline
\multicolumn{6}{|l|}{Total} \\
\hline
25 classes & 12716 & 10 classes & 1506 & 20 classes & 3377 \\
\hline
\end{tabular}}
\end{center}
\caption{Split composition of ShapeNet55-LS }
\label{tbl:shapenet-split}
\end{table}

\begin{table}[b!]
\begin{center}
\setlength{\tabcolsep}{3.5pt}
\scalebox{0.80}{
\begin{tabular}{|lr|lr|lr|}
\hline
Training & \# samples & Validation & \# samples  & Testing & \# samples \\
\hline

bed           & 615 & cup        & 99  & range hood & 215 \\
car           & 297 & xbox       & 123 & bowl       & 84  \\
guitar        & 255 & bathtub    & 156 & stool      & 110 \\
bottle        & 435 & cone       & 187 & radio      & 124 \\
desk          & 286 & curtain    & 158 & stairs     & 144 \\
night stand   & 286 & door       & 129 & lamp       & 144 \\
glass box     & 271 & flower pot & 169 & tent       & 183 \\
sofa          & 780 & person     & 108 & sink       & 148 \\
piano         & 331 & wardrobe   & 107 & bench      & 193 \\
toilet        & 444 & keyboard   & 165 & laptop     & 169 \\
monitor       & 565 &            &     &            &     \\
table         & 492 &            &     &            &     \\
dresser       & 286 &            &     &            &     \\
airplane      & 726 &            &     &            &     \\
tv stand      & 367 &            &     &            &     \\
chair         & 989 &            &     &            &     \\
bookshelf     & 672 &            &     &            &     \\
vase          & 575 &            &     &            &     \\
plant         & 340 &            &     &            &     \\
mantel        & 384 &            &     &            &     \\

\hline
\multicolumn{6}{|l|}{Total} \\
\hline
20 classes & 9396 & 10 classes & 1401 & 10 classes & 1514 \\
\hline
\end{tabular}}
\end{center}
\caption{Split composition of ModelNet40-LS}
\label{tbl:modelnet-split}
\end{table}

\begin{table}[t!]
\begin{center}
\setlength{\tabcolsep}{3.5pt}
\scalebox{0.80}{
\begin{tabular}{|lr|lr|lr|}
\hline
Training & \# samples & Validation & \# samples  & Testing & \# samples \\
\hline

candy        & 56  & airplane  & 35 & boat           & 38 \\
flower       & 54  & shark     & 30 & lion           & 17 \\
dragon       & 43  & truck     & 34 & whale          & 41 \\
apple        & 54  & phone     & 23 & cupcake        & 28 \\
guitar       & 55  & giraffe   & 15 & train          & 22 \\
tree         & 57  & horse     & 37 & pizza          & 26 \\
glass        & 63  & fish      & 37 & marker         & 19 \\
cup          & 60  & fan       & 31 & cookie         & 28 \\
pig          & 41  & shoe      & 41 & sandwich       & 15 \\
cat          & 79  & snake     & 32 & octopus        & 31 \\
chair        & 210 &           &    & monkey         & 16 \\
ice cream    & 43  &           &    & fries          & 15 \\
hat          & 64  &           &    & violin         & 25 \\
deer moose   & 65  &           &    & mushroom       & 23 \\
penguin      & 53  &           &    & closet         & 15 \\
ball         & 44  &           &    & tractor        & 16 \\
fox          & 64  &           &    & submarine      & 18 \\
dog          & 103 &           &    & butterfly      & 18 \\
knife        & 45  &           &    & pear           & 18 \\
laptop       & 41  &           &    & bicycle        & 17 \\
pen          & 42  &           &    & dolphin        & 25 \\
mug          & 97  &           &    & bunny          & 27 \\
plate        & 50  &           &    & coin           & 33 \\
chess piece  & 49  &           &    & radio          & 40 \\
cake         & 48  &           &    & grapes         & 16 \\
frog         & 43  &           &    & banana         & 35 \\
ladder       & 53  &           &    & cow            & 25 \\
keyboard     & 51  &           &    & donut          & 34 \\
sofa         & 63  &           &    & stove          & 29 \\
trashcan     & 44  &           &    & sink           & 25 \\
dinosaur     & 76  &           &    & orange         & 24 \\
bottle       & 111 &           &    & saw            & 19 \\
elephant     & 46  &           &    & chicken        & 25 \\
pencil       & 50  &           &    & hamburger      & 16 \\
key          & 49  &           &    & piano          & 39 \\
monitor      & 57  &           &    & light bulb     & 15 \\
hammer       & 94  &           &    & spade          & 36 \\
screwdriver  & 46  &           &    & crab           & 40 \\
robot        & 105 &           &    & sheep          & 40 \\
bread        & 38  &           &    & toaster        & 21 \\
             &     &           &    & lizard         & 20 \\
             &     &           &    & motorcycle     & 16 \\
             &     &           &    & mouse          & 25 \\
             &     &           &    & pc mouse       & 15 \\
             &     &           &    & bus            & 18 \\
             &     &           &    & helicopter     & 20 \\
             &     &           &    & microwave      & 18 \\
             &     &           &    & cells battery  & 41 \\
             &     &           &    & drum           & 26 \\
             &     &           &    & panda          & 24 \\
             &     &           &    & tv             & 21 \\
             &     &           &    & car            & 28 \\
             &     &           &    & helmet         & 17 \\
             &     &           &    & fridge         & 31 \\
             &     &           &    & bowl           & 28 \\

\hline
\multicolumn{6}{|l|}{Total} \\
\hline
40 classes & 2506 & 10 classes & 315 & 55 classes & 1358\\
\hline
\end{tabular}}
\end{center}
\caption{Split composition of Toys4K}
\label{tbl:toys-split}
\end{table}

%% file: main.bbl
\begin{thebibliography}{10}\itemsep=-1pt

\bibitem{simpleshot-repo}
\url{https://github.com/mileyan/simple_shot}.

\bibitem{RFS-repo}
\url{https://github.com/WangYueFt/rfs/}.

\bibitem{FEAT-repo}
\url{https://github.com/Sha-Lab/FEAT}.

\bibitem{pointnet-repo}
\url{https://github.com/yanx27/Pointnet_Pointnet2_pytorch}.

\bibitem{dgcnn-repo}
\url{https://github.com/AnTao97/dgcnn.pytorch}.

\bibitem{chen2018closer}
Wei-Yu Chen, Yen-Cheng Liu, Zsolt Kira, Yu-Chiang~Frank Wang, and Jia-Bin
  Huang.
\newblock A closer look at few-shot classification.
\newblock In {\em International Conference on Learning Representations}, 2018.

\bibitem{he2016deep1}
Kaiming He, Xiangyu Zhang, Shaoqing Ren, and Jian Sun.
\newblock Deep residual learning for image recognition.
\newblock In {\em Proceedings of the IEEE conference on computer vision and
  pattern recognition}, pages 770--778, 2016.

\bibitem{kingma2014adam}
Diederik~P Kingma and Jimmy Ba.
\newblock Adam: A method for stochastic optimization.
\newblock {\em arXiv preprint arXiv:1412.6980}, 2014.

\bibitem{paszke2019pytorch}
Adam Paszke, Sam Gross, Francisco Massa, Adam Lerer, James Bradbury, Gregory
  Chanan, Trevor Killeen, Zeming Lin, Natalia Gimelshein, Luca Antiga, et~al.
\newblock Pytorch: An imperative style, high-performance deep learning library.
\newblock In {\em Advances in neural information processing systems}, pages
  8026--8037, 2019.

\bibitem{scikit-learn}
F. Pedregosa, G. Varoquaux, A. Gramfort, V. Michel, B. Thirion, O. Grisel, M.
  Blondel, P. Prettenhofer, R. Weiss, V. Dubourg, J. Vanderplas, A. Passos, D.
  Cournapeau, M. Brucher, M. Perrot, and E. Duchesnay.
\newblock Scikit-learn: Machine learning in {P}ython.
\newblock {\em Journal of Machine Learning Research}, 12:2825--2830, 2011.

\bibitem{blender1}
Blender Proejct.
\newblock \url{https://blender.org}.

\bibitem{qi2017pointnet1}
Charles~R Qi, Hao Su, Kaichun Mo, and Leonidas~J Guibas.
\newblock Pointnet: Deep learning on point sets for 3d classification and
  segmentation.
\newblock In {\em Proceedings of the IEEE conference on computer vision and
  pattern recognition}, pages 652--660, 2017.

\bibitem{qi2017pointnet++1}
Charles~Ruizhongtai Qi, Li Yi, Hao Su, and Leonidas~J Guibas.
\newblock Pointnet++: Deep hierarchical feature learning on point sets in a
  metric space.
\newblock In {\em Advances in neural information processing systems}, pages
  5099--5108, 2017.

\bibitem{tian2020rethink1}
Yonglong Tian, Yue Wang, Dilip Krishnan, Joshua~B Tenenbaum, and Phillip Isola.
\newblock Rethinking few-shot image classification: a good embedding is all you
  need?
\newblock In {\em European Conference on Computer Vision (ECCV) 2020}, August
  2020.

\bibitem{vinyals2016matching1}
Oriol Vinyals, Charles Blundell, Timothy Lillicrap, Daan Wierstra, et~al.
\newblock Matching networks for one shot learning.
\newblock In {\em Advances in neural information processing systems}, pages
  3630--3638, 2016.

\bibitem{wang2019simpleshot1}
Yan Wang, Wei-Lun Chao, Kilian~Q Weinberger, and Laurens van~der Maaten.
\newblock Simpleshot: Revisiting nearest-neighbor classification for few-shot
  learning.
\newblock {\em arXiv preprint arXiv:1911.04623}, 2019.

\bibitem{wang2019dynamic1}
Yue Wang, Yongbin Sun, Ziwei Liu, Sanjay~E Sarma, Michael~M Bronstein, and
  Justin~M Solomon.
\newblock Dynamic graph cnn for learning on point clouds.
\newblock {\em Acm Transactions On Graphics (tog)}, 38(5):1--12, 2019.

\bibitem{pointnet2-repo}
Erik Wijmans.
\newblock Pointnet++ pytorch.
\newblock \url{https://github.com/erikwijmans/Pointnet2_PyTorch}, 2018.

\bibitem{Ye_2020_CVPR1}
Han-Jia Ye, Hexiang Hu, De-Chuan Zhan, and Fei Sha.
\newblock Few-shot learning via embedding adaptation with set-to-set functions.
\newblock In {\em Proceedings of the IEEE/CVF Conference on Computer Vision and
  Pattern Recognition (CVPR)}, June 2020.

\end{thebibliography}


\begin{thebibliography}{10}\itemsep=-1pt

\bibitem{blendswap}
blendswap.com.
\newblock \url{https://blendswap.com}.

\bibitem{chang2015shapenet}
Angel~X Chang, Thomas Funkhouser, Leonidas Guibas, Pat Hanrahan, Qixing Huang,
  Zimo Li, Silvio Savarese, Manolis Savva, Shuran Song, Hao Su, et~al.
\newblock Shapenet: An information-rich 3d model repository.
\newblock {\em arXiv preprint arXiv:1512.03012}, 2015.

\bibitem{chechik2009large}
Gal Chechik, Varun Sharma, Uri Shalit, and Samy Bengio.
\newblock Large scale online learning of image similarity through ranking.
\newblock In {\em Iberian Conference on Pattern Recognition and Image
  Analysis}, pages 11--14. Springer, 2009.

\bibitem{chenpointmixup}
Yunlu Chen, Vincent~Tao Hu, Efstratios Gavves, Thomas Mensink, Pascal Mettes,
  Pengwan Yang, and Cees~GM Snoek.
\newblock Pointmixup: Augmentation for point clouds.
\newblock In {\em Proceedings of the European Conference on Computer Vision
  (ECCV) 2020}.

\bibitem{deitke2020robothor}
Matt Deitke, Winson Han, Alvaro Herrasti, Aniruddha Kembhavi, Eric Kolve,
  Roozbeh Mottaghi, Jordi Salvador, Dustin Schwenk, Eli VanderBilt, Matthew
  Wallingford, et~al.
\newblock Robothor: An open simulation-to-real embodied ai platform.
\newblock In {\em Proceedings of the IEEE/CVF Conference on Computer Vision and
  Pattern Recognition}, pages 3164--3174, 2020.

\bibitem{deng2009imagenet}
Jia Deng, Wei Dong, Richard Socher, Li-Jia Li, Kai Li, and Li Fei-Fei.
\newblock Imagenet: A large-scale hierarchical image database.
\newblock In {\em 2009 IEEE conference on computer vision and pattern
  recognition}, pages 248--255. Ieee, 2009.

\bibitem{diesendruck2003specific}
Gil Diesendruck and Paul Bloom.
\newblock How specific is the shape bias?
\newblock {\em Child development}, 74(1):168--178, 2003.

\bibitem{dosovitskiy2017carla}
Alexey Dosovitskiy, German Ros, Felipe Codevilla, Antonio Lopez, and Vladlen
  Koltun.
\newblock Carla: An open urban driving simulator.
\newblock In {\em Conference on robot learning}, pages 1--16. PMLR, 2017.

\bibitem{edelman1999representation}
Shimon Edelman.
\newblock {\em Representation and recognition in vision}.
\newblock 1999.

\bibitem{feng2018gvcnn}
Yifan Feng, Zizhao Zhang, Xibin Zhao, Rongrong Ji, and Yue Gao.
\newblock Gvcnn: Group-view convolutional neural networks for 3d shape
  recognition.
\newblock In {\em Proceedings of the IEEE Conference on Computer Vision and
  Pattern Recognition}, pages 264--272, 2018.

\bibitem{finn2017model}
Chelsea Finn, Pieter Abbeel, and Sergey Levine.
\newblock Model-agnostic meta-learning for fast adaptation of deep networks.
\newblock In {\em Proceedings of the 34th International Conference on Machine
  Learning-Volume 70}, pages 1126--1135, 2017.

\bibitem{finn2018probabilistic}
Chelsea Finn, Kelvin Xu, and Sergey Levine.
\newblock Probabilistic model-agnostic meta-learning.
\newblock In {\em Advances in Neural Information Processing Systems}, pages
  9516--9527, 2018.

\bibitem{frome2013devise}
Andrea Frome, Greg~S Corrado, Jon Shlens, Samy Bengio, Jeff Dean, Marc'Aurelio
  Ranzato, and Tomas Mikolov.
\newblock Devise: A deep visual-semantic embedding model.
\newblock In {\em Advances in neural information processing systems}, pages
  2121--2129, 2013.

\bibitem{geirhos2018imagenet}
Robert Geirhos, Patricia Rubisch, Claudio Michaelis, Matthias Bethge, Felix~A
  Wichmann, and Wieland Brendel.
\newblock Imagenet-trained cnns are biased towards texture; increasing shape
  bias improves accuracy and robustness.
\newblock In {\em International Conference on Learning Representations}, 2018.

\bibitem{gershkoff2004shape}
Lisa Gershkoff-Stowe and Linda~B Smith.
\newblock Shape and the first hundred nouns.
\newblock {\em Child development}, 75(4):1098--1114, 2004.

\bibitem{he2016deep}
Kaiming He, Xiangyu Zhang, Shaoqing Ren, and Jian Sun.
\newblock Deep residual learning for image recognition.
\newblock In {\em Proceedings of the IEEE conference on computer vision and
  pattern recognition}, pages 770--778, 2016.

\bibitem{ho2020exploit}
Chih-Hui Ho, Bo Liu, Tz-Ying Wu, and Nuno Vasconcelos.
\newblock Exploit clues from views: Self-supervised and regularized learning
  for multiview object recognition.
\newblock In {\em Proceedings of the IEEE/CVF Conference on Computer Vision and
  Pattern Recognition}, pages 9090--9100, 2020.

\bibitem{hu2018relation}
Han Hu, Jiayuan Gu, Zheng Zhang, Jifeng Dai, and Yichen Wei.
\newblock Relation networks for object detection.
\newblock In {\em Proceedings of the IEEE Conference on Computer Vision and
  Pattern Recognition}, pages 3588--3597, 2018.

\bibitem{hubert2017learning}
Yao-Hung Hubert~Tsai, Liang-Kang Huang, and Ruslan Salakhutdinov.
\newblock Learning robust visual-semantic embeddings.
\newblock In {\em Proceedings of the IEEE International Conference on Computer
  Vision}, pages 3571--3580, 2017.

\bibitem{johnson2017clevr}
Justin Johnson, Bharath Hariharan, Laurens Van Der~Maaten, Li Fei-Fei, C
  Lawrence~Zitnick, and Ross Girshick.
\newblock Clevr: A diagnostic dataset for compositional language and elementary
  visual reasoning.
\newblock In {\em Proceedings of the IEEE Conference on Computer Vision and
  Pattern Recognition}, pages 2901--2910, 2017.

\bibitem{koch2019abc}
Sebastian Koch, Albert Matveev, Zhongshi Jiang, Francis Williams, Alexey
  Artemov, Evgeny Burnaev, Marc Alexa, Denis Zorin, and Daniele Panozzo.
\newblock Abc: A big cad model dataset for geometric deep learning.
\newblock In {\em Proceedings of the IEEE Conference on Computer Vision and
  Pattern Recognition}, pages 9601--9611, 2019.

\bibitem{krizhevsky2009learning}
Alex Krizhevsky, Geoffrey Hinton, et~al.
\newblock Learning multiple layers of features from tiny images.
\newblock 2009.

\bibitem{landau1998object}
Barbara Landau, Linda Smith, and Susan Jones.
\newblock Object shape, object function, and object name.
\newblock {\em Journal of memory and language}, 38(1):1--27, 1998.

\bibitem{landau1988importance}
Barbara Landau, Linda~B Smith, and Susan~S Jones.
\newblock The importance of shape in early lexical learning.
\newblock {\em Cognitive development}, 3(3):299--321, 1988.

\bibitem{lee2018cross}
Tang Lee, Yen-Liang Lin, HungYueh Chiang, Ming-Wei Chiu, Winston Hsu, and Polly
  Huang.
\newblock Cross-domain image-based 3d shape retrieval by view sequence
  learning.
\newblock In {\em 2018 International Conference on 3D Vision (3DV)}, pages
  258--266. IEEE, 2018.

\bibitem{li2015joint}
Yangyan Li, Hao Su, Charles~Ruizhongtai Qi, Noa Fish, Daniel Cohen-Or, and
  Leonidas~J Guibas.
\newblock Joint embeddings of shapes and images via cnn image purification.
\newblock {\em ACM transactions on graphics (TOG)}, 34(6):1--12, 2015.

\bibitem{marr2010vision}
David Marr.
\newblock {\em Vision: A computational investigation into the human
  representation and processing of visual information}.
\newblock W.H. Freeman and Company, 1982.

\bibitem{monti2017geometric}
Federico Monti, Davide Boscaini, Jonathan Masci, Emanuele Rodola, Jan Svoboda,
  and Michael~M Bronstein.
\newblock Geometric deep learning on graphs and manifolds using mixture model
  cnns.
\newblock In {\em Proceedings of the IEEE Conference on Computer Vision and
  Pattern Recognition}, pages 5115--5124, 2017.

\bibitem{Mundy2006}
Joseph~L. Mundy.
\newblock {Object Recognition in the Geometric Era: A Retrospective}.
\newblock In Jean Ponce, Martial Hebert, Cordelia Schmid, and Andrew Zisserman,
  editors, {\em Toward Category Level Object Recognition}, pages 3--28.
  Springer, 2006.

\bibitem{nichol2018reptile}
Alex Nichol and John Schulman.
\newblock Reptile: a scalable metalearning algorithm.
\newblock {\em arXiv preprint arXiv:1803.02999}, 2(3):4, 2018.

\bibitem{osada2002shape}
Robert Osada, Thomas Funkhouser, Bernard Chazelle, and David Dobkin.
\newblock Shape distributions.
\newblock {\em ACM Transactions on Graphics (TOG)}, 21(4):807--832, 2002.

\bibitem{poly}
poly.google.com/.
\newblock \url{https://poly.google.com/}.

\bibitem{blender}
Blender Proejct.
\newblock \url{https://blender.org}.

\bibitem{qi2017pointnet}
Charles~R Qi, Hao Su, Kaichun Mo, and Leonidas~J Guibas.
\newblock Pointnet: Deep learning on point sets for 3d classification and
  segmentation.
\newblock In {\em Proceedings of the IEEE conference on computer vision and
  pattern recognition}, pages 652--660, 2017.

\bibitem{qi2016volumetric}
Charles~R Qi, Hao Su, Matthias Nie{\ss}ner, Angela Dai, Mengyuan Yan, and
  Leonidas~J Guibas.
\newblock Volumetric and multi-view cnns for object classification on 3d data.
\newblock In {\em Proceedings of the IEEE conference on computer vision and
  pattern recognition}, pages 5648--5656, 2016.

\bibitem{qi2017pointnet++}
Charles~Ruizhongtai Qi, Li Yi, Hao Su, and Leonidas~J Guibas.
\newblock Pointnet++: Deep hierarchical feature learning on point sets in a
  metric space.
\newblock In {\em Advances in neural information processing systems}, pages
  5099--5108, 2017.

\bibitem{ren2018meta}
Mengye Ren, Eleni Triantafillou, Sachin Ravi, Jake Snell, Kevin Swersky,
  Joshua~B Tenenbaum, Hugo Larochelle, and Richard~S Zemel.
\newblock Meta-learning for semi-supervised few-shot classification.
\newblock {\em arXiv preprint arXiv:1803.00676}, 2018.

\bibitem{rusu2018meta}
Andrei~A Rusu, Dushyant Rao, Jakub Sygnowski, Oriol Vinyals, Razvan Pascanu,
  Simon Osindero, and Raia Hadsell.
\newblock Meta-learning with latent embedding optimization.
\newblock {\em arXiv preprint arXiv:1807.05960}, 2018.

\bibitem{schonfeld2019generalized}
Edgar Schonfeld, Sayna Ebrahimi, Samarth Sinha, Trevor Darrell, and Zeynep
  Akata.
\newblock Generalized zero-and few-shot learning via aligned variational
  autoencoders.
\newblock In {\em Proceedings of the IEEE Conference on Computer Vision and
  Pattern Recognition}, pages 8247--8255, 2019.

\bibitem{schroff2015facenet}
Florian Schroff, Dmitry Kalenichenko, and James Philbin.
\newblock Facenet: A unified embedding for face recognition and clustering.
\newblock In {\em Proceedings of the IEEE conference on computer vision and
  pattern recognition}, pages 815--823, 2015.

\bibitem{schwartz2019baby}
Eli Schwartz, Leonid Karlinsky, Rogerio Feris, Raja Giryes, and Alex~M
  Bronstein.
\newblock Baby steps towards few-shot learning with multiple semantics.
\newblock {\em arXiv preprint arXiv:1906.01905}, 2019.

\bibitem{shilane2004princeton}
Philip Shilane, Patrick Min, Michael Kazhdan, and Thomas Funkhouser.
\newblock The princeton shape benchmark.
\newblock In {\em Proceedings Shape Modeling Applications, 2004.}, pages
  167--178. IEEE, 2004.

\bibitem{sketchfab}
sketchfab.com.
\newblock \url{https://sketchfab.com}.

\bibitem{snell2017prototypical}
Jake Snell, Kevin Swersky, and Richard Zemel.
\newblock Prototypical networks for few-shot learning.
\newblock In {\em Advances in neural information processing systems}, pages
  4077--4087, 2017.

\bibitem{stojanov2019incremental}
Stefan Stojanov, Samarth Mishra, Ngoc~Anh Thai, Nikhil Dhanda, Ahmad Humayun,
  Chen Yu, Linda~B Smith, and James~M Rehg.
\newblock Incremental object learning from contiguous views.
\newblock In {\em Proceedings of the IEEE Conference on Computer Vision and
  Pattern Recognition}, pages 8777--8786, 2019.

\bibitem{su2015multi}
Hang Su, Subhransu Maji, Evangelos Kalogerakis, and Erik Learned-Miller.
\newblock Multi-view convolutional neural networks for 3d shape recognition.
\newblock In {\em Proceedings of the IEEE international conference on computer
  vision}, pages 945--953, 2015.

\bibitem{sung2018learning}
Flood Sung, Yongxin Yang, Li Zhang, Tao Xiang, Philip~HS Torr, and Timothy~M
  Hospedales.
\newblock Learning to compare: Relation network for few-shot learning.
\newblock In {\em Proceedings of the IEEE Conference on Computer Vision and
  Pattern Recognition}, pages 1199--1208, 2018.

\bibitem{tatsuma2012large}
Atsushi Tatsuma, Hitoshi Koyanagi, and Masaki Aono.
\newblock A large-scale shape benchmark for 3d object retrieval: Toyohashi
  shape benchmark.
\newblock In {\em Proceedings of The 2012 Asia Pacific Signal and Information
  Processing Association Annual Summit and Conference}, pages 1--10. IEEE,
  2012.

\bibitem{tian2020rethink}
Yonglong Tian, Yue Wang, Dilip Krishnan, Joshua~B Tenenbaum, and Phillip Isola.
\newblock Rethinking few-shot image classification: a good embedding is all you
  need?
\newblock In {\em European Conference on Computer Vision (ECCV) 2020}, August
  2020.

\bibitem{turbosquid}
turbosquid.com.
\newblock \url{https://turbosquid.com}.

\bibitem{Ullman1996}
Shimon Ullman.
\newblock {\em High Level Vision: Object Recognition and Visual Cognition}.
\newblock MIT Press, 1996.

\bibitem{vaswani2017attention}
Ashish Vaswani, Noam Shazeer, Niki Parmar, Jakob Uszkoreit, Llion Jones,
  Aidan~N Gomez, {\L}ukasz Kaiser, and Illia Polosukhin.
\newblock Attention is all you need.
\newblock In {\em Advances in neural information processing systems}, pages
  5998--6008, 2017.

\bibitem{vinyals2016matching}
Oriol Vinyals, Charles Blundell, Timothy Lillicrap, Daan Wierstra, et~al.
\newblock Matching networks for one shot learning.
\newblock In {\em Advances in neural information processing systems}, pages
  3630--3638, 2016.

\bibitem{wang2019simpleshot}
Yan Wang, Wei-Lun Chao, Kilian~Q Weinberger, and Laurens van~der Maaten.
\newblock Simpleshot: Revisiting nearest-neighbor classification for few-shot
  learning.
\newblock {\em arXiv preprint arXiv:1911.04623}, 2019.

\bibitem{wang2019dynamic}
Yue Wang, Yongbin Sun, Ziwei Liu, Sanjay~E Sarma, Michael~M Bronstein, and
  Justin~M Solomon.
\newblock Dynamic graph cnn for learning on point clouds.
\newblock {\em Acm Transactions On Graphics (tog)}, 38(5):1--12, 2019.

\bibitem{wu20153d}
Zhirong Wu, Shuran Song, Aditya Khosla, Fisher Yu, Linguang Zhang, Xiaoou Tang,
  and Jianxiong Xiao.
\newblock 3d shapenets: A deep representation for volumetric shapes.
\newblock In {\em Proceedings of the IEEE conference on computer vision and
  pattern recognition}, pages 1912--1920, 2015.

\bibitem{xian2018feature}
Yongqin Xian, Tobias Lorenz, Bernt Schiele, and Zeynep Akata.
\newblock Feature generating networks for zero-shot learning.
\newblock In {\em Proceedings of the IEEE conference on computer vision and
  pattern recognition}, pages 5542--5551, 2018.

\bibitem{xing2019adaptive}
Chen Xing, Negar Rostamzadeh, Boris Oreshkin, and Pedro~OO Pinheiro.
\newblock Adaptive cross-modal few-shot learning.
\newblock In {\em Advances in Neural Information Processing Systems}, pages
  4847--4857, 2019.

\bibitem{Ye_2020_CVPR}
Han-Jia Ye, Hexiang Hu, De-Chuan Zhan, and Fei Sha.
\newblock Few-shot learning via embedding adaptation with set-to-set functions.
\newblock In {\em Proceedings of the IEEE/CVF Conference on Computer Vision and
  Pattern Recognition (CVPR)}, June 2020.

\bibitem{zhou2016thingi10k}
Qingnan Zhou and Alec Jacobson.
\newblock Thingi10k: A dataset of 10,000 3d-printing models.
\newblock {\em arXiv preprint arXiv:1605.04797}, 2016.

\end{thebibliography}
